\documentclass[11pt]{article}

\usepackage[preprint]{acl}

\usepackage{times}
\usepackage{latexsym}

\usepackage[T1]{fontenc}

\usepackage[utf8]{inputenc}

\usepackage{microtype}

\usepackage{inconsolata}

\usepackage{graphicx}
\usepackage{subcaption}

\usepackage{amsmath}
\usepackage{amssymb}

\usepackage{multirow}
\usepackage{booktabs}
\usepackage{colortbl}
\usepackage{xcolor}

\usepackage[many]{tcolorbox}
\tcbuselibrary{listings, skins, breakable}

\newtcolorbox[auto counter, number within=section]{listingbox}[2][]{%
  enhanced,
  breakable,
  colback=gray!5,       
  colframe=black!30,    
  coltitle=black,       
  fonttitle=\bfseries,
  title=Listing~\thetcbcounter:~#2, 
  sharp corners,
  boxrule=0.8pt,
  left=6pt, right=6pt, top=6pt, bottom=6pt,
  before skip=10pt, after skip=10pt,
  #1
}

\usepackage{booktabs}
\usepackage{multirow}
\usepackage{colortbl}
\usepackage{siunitx}
\usepackage{makecell}
\usepackage{enumitem}
\usepackage{algorithm}
\usepackage{algpseudocode}

\title{Hallucination-R1: Robustness-Oriented Paraphrase Generation for Factual Consistency}

\author{Wenhan Yu\thanks{These authors contributed equally to this work.}\quad
  Wenxin Wu\textsuperscript{*}\quad
  Hao Wang\quad
  Lei Sha\thanks{Corresponding author.} \\
  Beihang University \\
  Beijing, 100191, China \\
  \texttt{yuwenhan@buaa.edu.cn}\quad
  \texttt{wwx20030520@buaa.edu.cn} \\
  \texttt{wanghao\_ai@buaa.edu.cn}\quad
  \texttt{shalei@buaa.edu.cn}}

\begin{document}
\maketitle
\begin{abstract}
Factual hallucination is commonly defined by incorrect factual outputs. We study a paraphrase-induced hallucination setting, where a model answers a factual question correctly in its original form but generates an incorrect answer under a semantically equivalent paraphrase. Such inconsistencies expose latent factual instability under semantic invariance. However, general-purpose paraphrases are often insufficient as robustness-oriented supervision: near-copy paraphrases provide weak signals, while overly diverse paraphrases may break semantic equivalence.
In this paper, we propose \textsc{Hallucination-R1}, a robustness-oriented paraphrase generation framework that learns to produce semantically faithful yet robustness-challenging paraphrases for factual consistency. Through two-stage optimization, it first stabilizes meaning-preserving and diverse paraphrasing, then rewards paraphrases that reveal factual consistency degradation in downstream QA models. Experiments on SimpleQuestions, PopQA, and TruthfulQA show that \textsc{Hallucination-R1} achieves a strong consistency--diversity trade-off and exposes robustness failures across multiple model families and datasets. Further analyses indicate that these failures are not reducible to surface-level artifacts or semantic drift, but reveal non-trivial factual instability under meaning-preserving variation. A lightweight fine-tuning study also shows that \textsc{Hallucination-R1}-generated data improves robust accuracy under paraphrase variations, suggesting its utility for robustness-oriented training.
Our code and models are publicly available at \url{https://github.com/yuwenhan07/Hallucination-R1}.
\end{abstract}

\section{Introduction}

Large language models (LLMs) have shown strong question-answering capabilities, yet their factual behavior can remain unstable under semantics-preserving paraphrases \cite{brown2020language,singh2025openai}. A model may answer an original factual question correctly, but produce an incorrect response when the same query is expressed in a different surface form \cite{ribeiro2018semantically,gan2019improving}. Since the factual intent remains unchanged, such failures suggest that hallucination is not merely caused by knowledge absence, but can also reflect unstable factual behavior, where the model's internal preference over known facts drifts under meaning-preserving linguistic variation \cite{lin2022truthfulqa,huang2025survey}.

This motivates a shift from treating paraphrases merely as diagnostic probes \cite{ribeiro2020beyond} to learning paraphrase distributions that can provide useful supervision for improving factual robustness. For robustness-oriented fine-tuning, such data should preserve the original factual intent while introducing meaningful linguistic variation, enabling models to learn stable factual behavior across different surface forms. This is difficult because near-copy paraphrases provide limited robustness signals, whereas overly aggressive paraphrases may change the underlying factual query~\cite{feng2021survey}.

In this paper, we propose \textsc{Hallucination-R1}, a robustness-oriented paraphrase generation framework for factual consistency. Instead of constructing arbitrary adversarial prompts, \textsc{Hallucination-R1} learns to generate semantically consistent yet robustness-challenging paraphrases for factual question answering. The framework follows a two-stage optimization pipeline: it first learns to generate semantically valid and diverse paraphrases, and then further optimizes them to expose factual consistency degradation in downstream QA models under semantic invariance.

Experiments on SimpleQuestions, PopQA, and TruthfulQA show that \textsc{Hallucination-R1} generates semantically consistent and diverse paraphrases that expose factual robustness failures across multiple QA models, including settings beyond the training dataset and feedback model. Further analyses suggest that these failures are non-trivial, and a lightweight fine-tuning study demonstrates the utility of the generated data for improving robust accuracy under paraphrase variations.

Our contributions are summarized as follows:
\begin{itemize}
    \item We formulate robustness-oriented paraphrase generation for factual consistency, where paraphrases must preserve semantics while exposing factual instability under surface-form variation.
    \item We propose \textsc{Hallucination-R1}, a two-stage GRPO-based framework that combines semantic consistency, group-relative diversity, and factual inconsistency exposure signals.
    \item We show that \textsc{Hallucination-R1} exposes factual robustness failures across multiple QA models and datasets, and that these failures reflect non-trivial factual instability rather than simple surface-level artifacts.
    \item We demonstrate the training utility of \textsc{Hallucination-R1}-generated paraphrases, as they improve robust accuracy under semantics-preserving paraphrase variations.
\end{itemize}
\section{Related Work}

\subsection{Factual Hallucination and Paraphrase Inconsistency}

Factual hallucination broadly refers to cases where large language models generate outputs that are inconsistent with real-world facts or unverifiable, including fabricated events, incorrect dates, or nonexistent references~\cite{huang2025survey,ji2023survey}. Prior studies attribute such errors to factors including biased or incomplete training data, outdated knowledge, and overconfident decoding behaviors~\cite{li2024dawn,ji2023survey}. Recent work further shows that language models may produce highly misleading factual errors even when semantic content and surface-level fluency are largely preserved, posing substantial risks to high-reliability applications~\cite{zhang2023language,lin2022truthfulqa}.

In contrast to settings where factual errors primarily indicate missing, outdated, or inaccessible knowledge, this work focuses on cases where the model answers the original question correctly but fails under a semantically equivalent paraphrase. Such failures suggest that the model may possess the relevant factual knowledge, yet fail to maintain a stable factual response under meaning-preserving linguistic variation \cite{manakul2023selfcheckgpt,choi2025roparq}. Since directly determining whether a model truly "knows" a fact is inherently difficult, inconsistencies across semantically equivalent paraphrases provide a practical signal for characterizing latent factual instability \cite{choi2025roparq}.

\subsection{Paraphrase Robustness Evaluation}

Recent studies have shown that language models are highly sensitive to semantics-preserving reformulations \cite{gan2019improving,choi2025roparq}. Even when the underlying factual intent remains unchanged, paraphrase variations can lead to substantial performance degradation or inconsistent predictions \cite{lunardi2025robustness}. Existing work mainly uses paraphrases as evaluation probes, measuring whether model predictions remain consistent across semantically equivalent question forms \cite{gan2019improving,choi2025roparq}. These studies reveal the brittleness of current language models under paraphrase variation.

However, existing paraphrase-based evaluation and training studies typically construct paraphrases through predefined paraphrasing strategies, similarity filters, or diversity-oriented criteria. Although recent work on worst-case prompting and latent adversarial robustness further highlights the importance of challenging input variations \cite{cao2024worst,fu2025same}, existing paraphrase construction largely focuses on semantic preservation and linguistic diversity, without explicitly targeting paraphrase patterns that expose factual robustness fragility. This leaves open the question of how to generate paraphrases that are not only semantically valid, but also effective for revealing factual instability and supporting robustness-oriented training.

\subsection{Robustness-Oriented Paraphrase Generation}

Existing paraphrase generation and text augmentation methods provide useful mechanisms for producing meaning-preserving or linguistically diverse paraphrases \cite{huang2023paraamr,feng2021survey}. Building on the gap above, we argue that factual robustness requires a more targeted paraphrase objective. A useful paraphrase should preserve the answer-bearing intent while exposing whether a QA model can maintain factual consistency across meaning-preserving variations.

We therefore formulate paraphrase generation as a robustness-oriented learning problem. \textsc{Hallucination-R1} learns to generate semantically constrained paraphrases that preserve factual intent while exposing factual consistency degradation in QA models, making them useful for both robustness evaluation and training.
\begin{figure*}
    \centering
    \includegraphics[width=0.95\linewidth]{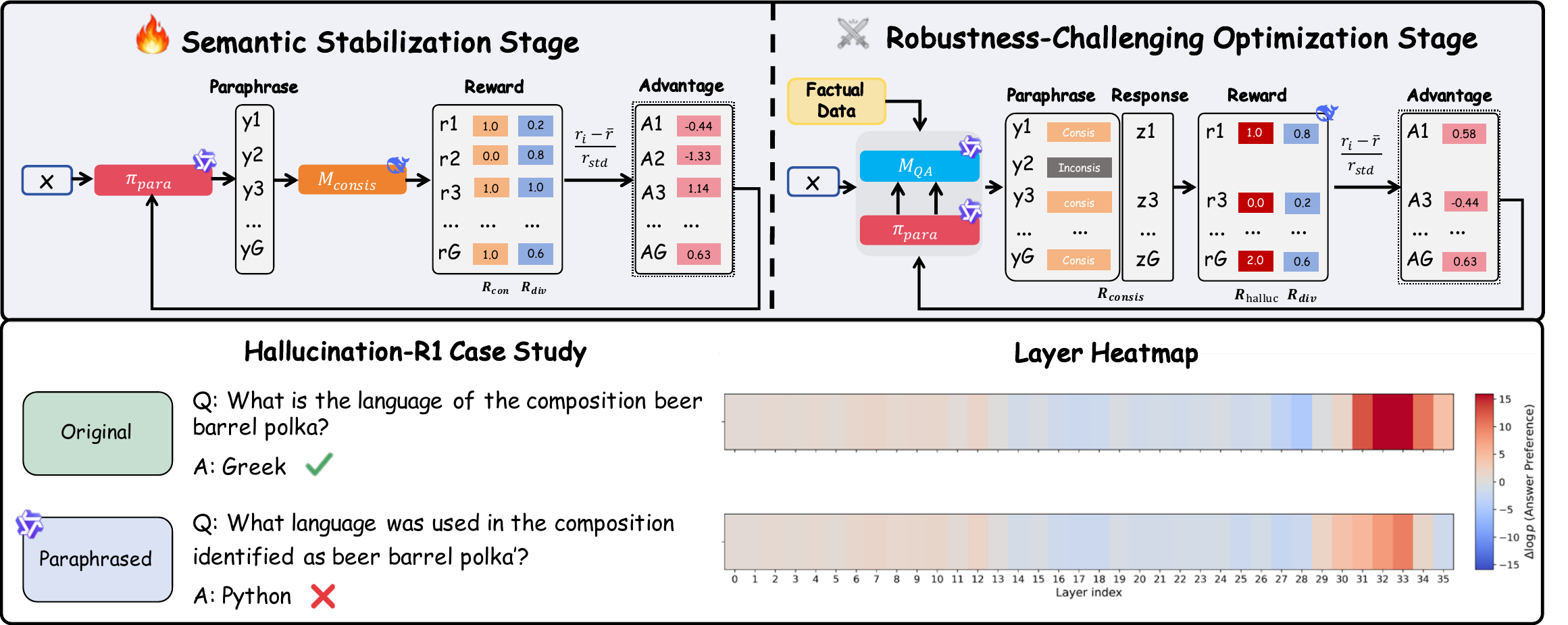}
    \caption{\textbf{Overview of the \textsc{Hallucination-R1} framework.}
The upper panel illustrates the two-stage training pipeline: the \textbf{semantic stabilization stage} learns semantically preserved paraphrases with controlled diversity exploration, while the \textbf{robustness-challenging optimization stage} optimizes paraphrases to expose factual inconsistency in a QA model under semantic invariance.
The lower panel shows a representative \textbf{case study}, where a semantically equivalent paraphrase shifts the QA model from a correct answer to a factual hallucination. The heatmap visualizes layer-wise log-probability preference changes, with red indicating stronger preference for the correct answer token and blue for the incorrect one.}
    \label{fig:pipeline}
\end{figure*}

\section{Methodology}
\label{sec:method}

We aim to generate robustness-oriented paraphrases for factual question answering.
A useful paraphrase should remain semantically equivalent to the original question, introduce diverse surface-form variations, and reveal whether a QA model can maintain factual consistency under meaning-preserving reformulations.
\textsc{Hallucination-R1} therefore treats paraphrase generation as a constrained robustness-oriented learning problem.

As shown in Figure~\ref{fig:pipeline}, we train a dedicated paraphrase generator through a two-stage optimization pipeline.
The first stage stabilizes semantic-preserving paraphrase generation with controlled diversity, while the second stage further encourages paraphrases that expose factual inconsistency in downstream QA models.

\subsection{Semantic-Preserving Challenging Paraphrase Generation}

Given an original factual question $x$, we train a paraphrase generator $\pi_{\theta}$ to produce a paraphrase $y \sim \pi_{\theta}(\cdot \mid x)$.
We decompose the optimization signal into three components: semantic consistency, paraphrase diversity, and factual inconsistency exposure.

\paragraph{Semantic consistency.}
Semantic consistency is enforced as a hard constraint in our framework.
We use an LLM-as-judge $M_{\text{consis}}$ to determine whether a paraphrase preserves the original meaning.
In our reward design, semantic consistency serves as a hard gate: if a paraphrase is judged semantically inconsistent, its overall reward is set to zero regardless of other signals.
We denote the corresponding consistency reward as $R_{\text{consis}}$.

\paragraph{Diversity.}
To prevent semantically consistent paraphrases from collapsing into near copies, we introduce a group-relative diversity reward $R_{\text{div}}$.
For each question, we sample a group of candidate paraphrases and assign rank-based rewards according to the lexical and semantic dissimilarity between each semantically consistent candidate and the others in the same group.
This group-relative scoring encourages exploration by rewarding candidates that are more diverse than their peers, rather than those that merely pass a fixed diversity threshold.
Details are provided in Appendix~\ref{app:warmup-stage}.

\paragraph{Factual inconsistency exposure.}
We further encourage the generator to produce paraphrases that reveal factual instability in downstream QA models.
For each paraphrase, the QA model first generates an answer $z$, and an LLM-as-judge $M_{\text{halluc}}$ evaluates whether $z$ is factually inconsistent with the ground-truth answer.
The judge outputs a factual inconsistency score $R_{\text{halluc}}\in\{0,1,2\}$, where higher values indicate stronger factual inconsistency.
Details are provided in Appendix~\ref{app:harmful-stage}.

\paragraph{Sequence-level optimization.}
Our objective naturally fits GRPO, since diversity is inherently a group-level property: it can only be measured across multiple paraphrases rather than from a single output. 
GRPO samples a group of candidate paraphrases and optimizes them with group-relative rewards, aligning well with our diversity reward while avoiding an explicit value function. 
The objective is:
\begin{equation}
\begin{aligned}
\max_{\theta}\;\;
&\mathbb{E}_{y \sim \pi_{\theta}(\cdot \mid x)}
    \left[ R(x, y, z) \right] \\
&-
\beta \, D_{\mathrm{KL}}\!\left(
    \pi_{\theta}(y \mid x)
    \,\|\,
    \pi_{\text{ref}}(y \mid x)
\right),
\end{aligned}
\end{equation}
where $z \sim \pi_{\text{QA}}(\cdot \mid y)$ denotes the answer generated by the QA model, $\pi_{\text{ref}}$ is the reference policy, and $\beta$ controls the strength of KL regularization.

\subsection{Two-Stage Training Pipeline}

Unless otherwise specified, \textsc{Hallucination-R1} is initialized from Qwen2.5-3B, and downstream QA feedback during training is obtained from Qwen2.5-3B as well. 
We use DeepSeek-V3 as the LLM judge for both semantic consistency verification and factual inconsistency scoring, corresponding to $M_{\text{consis}}$ and $M_{\text{halluc}}$, respectively.

\paragraph{Semantic Stabilization Stage.}
Before optimizing factual inconsistency exposure, we first stabilize semantic-preserving paraphrase generation, since the semantic-consistency hard gate would otherwise make downstream QA feedback highly sparse.
The first-stage reward is defined as:
\begin{equation}
R_{\text{stab}}
=
R_{\text{consis}}
+
\lambda_{\text{div}} R_{\text{div}},
\end{equation}
where $\lambda_{\text{div}}$ controls the strength of diversity exploration.

\paragraph{Robustness-Challenging Optimization Stage.}
After semantic stabilization, we further optimize the generator to expose factual inconsistency in downstream QA models. 
Given a generated paraphrase $y$ whose original question $x$ is correctly answered by the QA model, the QA model produces an answer $z$, and the LLM-as-judge $M_{\text{halluc}}$ evaluates whether $z$ is factually inconsistent with the ground-truth answer.
The training reward is:
\begin{equation}
R_{\text{train}} =
\begin{cases}
R_{\text{stab}} + R_{\text{halluc}}, & R_{\text{halluc}}\neq0, \\
0, & R_{\text{halluc}}=0,
\end{cases}
\end{equation}
Semantic consistency remains a hard constraint throughout training.
\section{Consistency--Diversity Trade-off Analysis}
\label{sec:consistency-diversity}

\begin{table*}[t]
\centering
\resizebox{\linewidth}{!}{
\begin{tabular}{l | c c c | c}
\toprule
\textbf{Paraphrase Model}
& \textbf{Consis.}
& \textbf{Mean Div.}
& \textbf{Ratio (Div. $>$ 0.9)}
& \textbf{Consis. $\times$ Mean Div.}   \\
\midrule
\midrule

Hallucination-R1-3B & 87.46\% & \textbf{0.930696} & \textbf{80.41\%} & 81.40\% \\
DeepSeek-V3         & 96.41\% & 0.880482 & 55.19\% & 84.89\% \\
Qwen2.5-3B          & 64.68\% & 0.900111 & 74.41\% & 58.22\% \\
Qwen2.5-32B         & 96.03\% & 0.892318 & 74.41\% & \textbf{85.69\%} \\
Qwen3-4B            & 98.28\% & 0.399141 & 3.16\% & 39.23\% \\
Qwen3-8B            & \textbf{98.71\%} & 0.47959 & 5.11\% & 47.34\% \\
Hallucination-R1 (Stage 1 Only)       & 80.37\% & 0.864836 & 58.56\% & 69.51\% \\

\bottomrule
\end{tabular}
}
\caption{
Consistency--diversity trade-off across different paraphrase models on the TruthfulQA dataset.
We report semantic consistency (Consis.), mean diversity (Mean Div.), consistency $\times$ diversity, and the ratio of high-diversity paraphrases (Div. $>$ 0.9).
Semantic consistency follows the definition used in the training reward.
}
\label{tab:tradeoff}
\end{table*}

We analyze the consistency--diversity trade-off across different paraphrasing models, including prompted LLM paraphrasing baselines. The consistency and diversity metrics follow the same definitions used in training rewards.

Table~\ref{tab:tradeoff} shows clear consistency--diversity differences across paraphrasing models. Some models achieve strong semantic consistency but produce limited expression variation, while others generate more diverse paraphrases at the cost of semantic stability. This imbalance is evident across model families: the Qwen3 series tends to produce conservative or near-copy paraphrases to maintain extremely high consistency, whereas Qwen2.5-3B is more flexible but often sacrifices semantic stability when increasing diversity. These patterns highlight the need for targeted optimization rather than relying on general-purpose paraphrasing.

In contrast, large models such as DeepSeek-V3 and Qwen2.5-32B already achieve strong consistency--diversity performance under prompting alone, as reflected by high Consis. $\times$ Mean Div. scores. \textsc{Hallucination-R1} achieves competitive results using only a 3B-scale model and produces substantially larger proportions of high-diversity paraphrases. This suggests that robustness-oriented paraphrase optimization can reshape the generation distribution of a smaller model toward semantically valid yet more varied paraphrases.

The comparison between the Stage-1-only model and the full \textsc{Hallucination-R1} further illustrates the role of the two-stage design. The Semantic Stabilization Stage improves the overall consistency--diversity balance mainly by increasing semantic consistency, despite a modest drop in diversity. After this stabilization, the robustness-challenging optimization stage further enhances diversity exploration under semantic-preserving constraints. This pattern indicates that semantic stabilization provides a reliable basis for later exploration, rather than simply encouraging arbitrary surface variation.

Taken together, these results consistently suggest that \textsc{Hallucination-R1} yields a favorable consistency--diversity trade-off for robustness-oriented paraphrase generation. While consistency and diversity capture basic paraphrase quality, the next section further evaluates whether the learned distribution is also effective at exposing downstream factual robustness failures.
\begin{table*}[t]
\centering
\resizebox{\linewidth}{!}{
\begin{tabular}{
l
| c c c
| c c c
| c c c
}
\toprule
\multirow{2}{*}{\textbf{QA Model}}
& \multicolumn{3}{c|}{\textbf{SimpleQuestions}}
& \multicolumn{3}{c|}{\textbf{PopQA}}
& \multicolumn{3}{c}{\textbf{TruthfulQA}} \\
\cmidrule(lr){2-4} \cmidrule(lr){5-7} \cmidrule(lr){8-10}

& \textbf{True (Acc)} & \textbf{ER@1} & \textbf{WER@5}
& \textbf{True (Acc)} & \textbf{ER@1} & \textbf{WER@5}
& \textbf{True (Acc)} & \textbf{ER@1} & \textbf{WER@5} \\
\midrule
\midrule
\multicolumn{10}{c}{\textbf{Hallucination-R1-1.5B}} \\
\midrule

Qwen2.5-3B
& 2106(21.25\%) & \textbf{30.44\%} & \textbf{71.37\%}
& 1702(11.93\%) & 30.14\% & 77.38\%
& 246(31.14\%) & 32.11\% & \textbf{83.33\%} \\

Qwen2.5-7B
& 2288(23.09\%) & 29.98\% & 66.35\%
& 2077(14.56\%) & 29.47\% & 73.57\%
& 323(40.89\%) & 30.96\% & 73.07\% \\

Qwen2.5-32B
& 2584(26.07\%) & 23.30\% & 56.93\%
& 2368(16.60\%) & 25.13\% & 66.17\%
& 374(47.34\%) & 24.60\% & 68.18\% \\

Qwen3-4B
& 2126(21.45\%) & 28.46\% & 68.30\%
& 1194(8.37\%) & \textbf{36.60\%} & \textbf{86.43\%}
& 233(29.49\%) & \textbf{36.91\%} & 81.55\% \\

LLaMA-3.1-8B
& 2489(25.12\%) & 23.66\% & 56.77\%
& 3082(21.60\%) & 22.97\% & 62.78\%
& 247(31.27\%) & 29.15\% & 67.61\% \\

DeepSeek-V3
& 3413(34.44\%) & 32.93\% & 46.35\%
& 4661(32.67\%) & 36.84\% & 48.27\%
& 391(49.49\%) & 22.51\% & 39.13\% \\

DeepSeek-R1
& 4258(42.97\%) & 24.10\% & 37.48\%
& 4560(31.96\%) & 43.75\% & 57.08\%
& 466(65.73\%) & 20.82\% & 33.48\% \\

GPT-5
& 5437(54.86\%) & 9.82\% & 25.31\% 
& 9912(69.48\%) & 7.31\% & 13.33\%
& 645(82.78\%) & 4.50\% & 11.94\% \\

\midrule
\midrule

\multicolumn{10}{c}{\textbf{Hallucination-R1-3B}} \\
\midrule

Qwen2.5-3B
& 2106(21.25\%) & \textbf{40.12\%} & \textbf{86.70\%}
& 1702(11.93\%) & 37.60\% & 82.20\%
& 246(31.14\%) & 47.97\% & 86.99\% \\

Qwen2.5-7B
& 2288(23.09\%) & 36.80\% & 73.73\%
& 2077(14.56\%) & 37.31\% & 81.51\%
& 323(40.89\%) & 45.51\% & 85.76\% \\

Qwen2.5-32B
& 2584(26.07\%) & 34.64\% & 71.17\%
& 2368(16.60\%) & 38.05\% & 81.88\%
& 374(47.34\%) & 48.40\% & 86.90\% \\

Qwen3-4B
& 2126(21.45\%) & 38.90\% & 75.68\%
& 1194(8.37\%) & 44.64\% & \textbf{90.28\%}
& 233(29.49\%) & 49.79\% & \textbf{87.12\%} \\

LLaMA-3.1-8B
& 2489(25.12\%) & 34.83\% & 71.11\%
& 3082(21.60\%) & 36.53\% & 80.89\%
& 247(31.27\%) & \textbf{55.06\%} & 87.04\% \\

DeepSeek-V3
& 3413(34.44\%) & 39.76\% & 51.36\%
& 4661(32.67\%) & 46.34\% & 60.78\%
& 391(49.49\%) & 41.69\% & 54.22\% \\

DeepSeek-R1
& 4258(42.97\%) & 38.14\% & 50.28\%
& 4560(31.96\%) & \textbf{49.76\%} & 65.37\%
& 466(65.73\%) & 41.85\% & 55.15\% \\

GPT-5
& 5437(54.86\%) & 9.91\% & 26.04\%
& 9912(69.48\%) & 8.34\% & 17.62\% 
& 645(82.78\%) & 6.20\% & 14.57\%  \\

\bottomrule
\end{tabular}
}
\caption{
Robustness evaluation results of \textsc{Hallucination-R1} across different QA models and benchmark datasets.
True (Acc) reports the number and percentage of samples that are correctly answered by the QA model before paraphrasing.
ER@1 and WER@5 denote Error Rate@1 and Worst-Case Error Rate@5, respectively.
A robustness failure is counted only when the rewritten question is judged to be semantically equivalent to the original input while causing the QA model to generate an incorrect answer.
All robustness metrics are computed over the subset of samples that are originally answered correctly by the QA model.
Results are reported for two \textsc{Hallucination-R1} scales: 1.5B and 3B parameters.
}
\label{tab:hallucination_results}
\end{table*}

\section{Robustness Evaluation}

We evaluate whether semantically equivalent yet challenging paraphrases generated by \textsc{Hallucination-R1} can expose factual consistency degradation across different QA models, architectures, and benchmark datasets.

We train \textsc{Hallucination-R1} on a filtered subset of SimpleQuestions with reliable factual correctness and evaluate it on SimpleQuestions, PopQA, and TruthfulQA. 
We consider two model scales, 1.5B and 3B, based on Qwen2.5-1.5B and Qwen2.5-3B, respectively, and train them using the two-stage pipeline described earlier. 
Full experimental details are provided in Appendix~\ref{app:experimental-setup}.

\subsection{Evaluation Protocol}

We evaluate robustness only on original questions that the QA model answers correctly. For each question, we generate $K$ paraphrases using $\pi_{\text{para}}$, verify semantic equivalence with $M_{\text{consis}}$, and evaluate the QA model's answer against the ground-truth answer using $M_{\text{judge}}$. A robustness failure is counted only when the paraphrase is semantically equivalent to the original question and causes the QA model to produce an incorrect answer. We set $K=1$ for ER@1 and $K=5$ for WER@5. The detailed procedure is provided in Appendix~\ref{app:robustness-evaluation-protocol}.

We report two complementary robustness metrics:

\begin{itemize}
    \item \textbf{Error Rate@1 (ER@1)}: the proportion of originally correct samples for which a single semantically equivalent paraphrase causes an incorrect answer.
    
    \item \textbf{Worst-Case Error Rate@5 (WER@5)}: the proportion of originally correct samples for which at least one of five generated paraphrases is both semantically equivalent and answered incorrectly.
\end{itemize}

\subsection{Robustness Evaluation Results}

Table~\ref{tab:hallucination_results} reports the robustness impact of paraphrases generated by \textsc{Hallucination-R1} under the strict constraint of semantic equivalence. Although the paraphrase generator is trained only on SimpleQuestions \cite{bordes2015large} with Qwen2.5-3B as the QA model, it consistently exposes factual consistency degradation across target models of different scales and architectures, as well as on distributionally distinct datasets such as PopQA \cite{mallen2023not} and TruthfulQA \cite{lin2022truthfulqa}.

Across evaluation settings, semantically equivalent paraphrases substantially reduce factual robustness under surface-form variation. This effect is pronounced in Worst-Case Error Rate@5, where many originally correct samples become incorrect under at least one of multiple meaning-preserving paraphrases. Moreover, the degradation remains observable even for larger and stronger models: while higher-capacity models generally show lower single-paraphrase error rates, they still exhibit non-negligible worst-case robustness failures. These results suggest that factual consistency under paraphrase variation remains a systematic challenge, and that \textsc{Hallucination-R1} provides effective robustness probes for evaluating factual stability under semantic-preserving transformations.

\subsection{Ablation Study}

\begin{table*}[htpb]
\centering
\resizebox{0.95\linewidth}{!}{
\begin{tabular}{l c | l l c c l l}
\toprule
\textbf{QA Model} 
& \textbf{Ori. Acc}
& \textbf{Paraphrase Model}
& \textbf{Consis@1}
& \textbf{RF@1}
& \textbf{RF@5}
& \textbf{ER@1}
& \textbf{WER@5} \\
\midrule
\midrule
\multirow{3}{*}{LLaMA-3.1-8B}
& \multirow{3}{*}{247 (31.27\%)}
& Qwen2.5-3B
& 64.68\% & 84  & 182 & 34.01\% & 73.68\% \\
& 
& Hallucination-R1 (Stage 1 Only)
& 80.37\% & 97  & 187 & 39.27\% & 75.71\% \\
& 
& \cellcolor{gray!20}\textbf{Hallucination-R1}
& \cellcolor{gray!20}\textbf{87.46\%$^{\textcolor{blue}{\uparrow 22.78\%}}$}
& \cellcolor{gray!20}\textbf{136}
& \cellcolor{gray!20}\textbf{215}
& \cellcolor{gray!20}\textbf{55.06\%$^{\textcolor{blue}{\uparrow 21.05\%}}$}
& \cellcolor{gray!20}\textbf{87.04\%$^{\textcolor{blue}{\uparrow 13.36\%}}$} \\
\midrule

\multirow{3}{*}{Qwen2.5-3B}
& \multirow{3}{*}{246 (31.14\%)}
& Qwen2.5-3B
& 64.68\% & 81  & 183 & 32.93\% & 74.39\% \\
& 
& Hallucination-R1 (Stage 1 Only)
& 80.37\% & 99  & 193 & 40.24\% & 78.46\% \\
& 
& \cellcolor{gray!20}\textbf{Hallucination-R1}
& \cellcolor{gray!20}\textbf{87.46\%$^{\textcolor{blue}{\uparrow 22.78\%}}$}
& \cellcolor{gray!20}\textbf{118}
& \cellcolor{gray!20}\textbf{214}
& \cellcolor{gray!20}\textbf{47.97\%$^{\textcolor{blue}{\uparrow 15.04\%}}$}
& \cellcolor{gray!20}\textbf{86.99\%$^{\textcolor{blue}{\uparrow 22.78\%}}$} \\

\midrule

\multirow{3}{*}{Qwen2.5-7B}
& \multirow{3}{*}{323 (40.89\%)}
& Qwen2.5-3B
& 64.68\% & 98  & 233 & 30.34\% & 72.14\% \\
& 
& Hallucination-R1 (Stage 1 Only)
& 80.37\% & 125 & 245 & 38.70\% & 75.85\% \\
& 
& \cellcolor{gray!20}\textbf{Hallucination-R1}
& \cellcolor{gray!20}\textbf{87.46\%$^{\textcolor{blue}{\uparrow 12.60\%}}$}
& \cellcolor{gray!20}\textbf{147}
& \cellcolor{gray!20}\textbf{277}
& \cellcolor{gray!20}\textbf{45.51\%$^{\textcolor{blue}{\uparrow 15.17\%}}$}
& \cellcolor{gray!20}\textbf{85.76\%$^{\textcolor{blue}{\uparrow 13.62\%}}$} \\
 

\bottomrule
\end{tabular}
}
\caption{
Ablation results of different paraphrase model variants across QA models on the TruthfulQA dataset. RF@1 denotes the number of robustness failures under a single semantically equivalent paraphrase, while RF@5 denotes the number of samples for which at least one out of five semantically equivalent paraphrases causes a robustness failure. $\uparrow$ / $\downarrow$ denote the absolute difference (in percentage points) of each variant relative to the full \textsc{Hallucination-R1-3B}, computed under the same QA model.
}
\label{tab:ablation}
\end{table*}

To analyze the contribution of different training stages in \textsc{Hallucination-R1}, we conduct ablation experiments on the TruthfulQA benchmark, with results summarized in Table~\ref{tab:ablation}. We compare the full \textsc{Hallucination-R1} model with two ablated variants: a model trained only with the Semantic Stabilization Stage, denoted as \textsc{Hallucination-R1} (Stage 1 Only), and a base paraphrasing model without reinforcement learning optimization (Qwen2.5-3B).

The results show that the Semantic Stabilization Stage provides an important foundation for generating semantically consistent paraphrases, while the subsequent Robustness-Challenging Optimization Stage substantially improves the effectiveness of paraphrases in exposing factual inconsistency under semantic-preserving transformations.

In addition, we compare the trained paraphrase generator with larger but untrained paraphrasing models such as Qwen2.5-7B. Despite its smaller scale, \textsc{Hallucination-R1-3B} consistently produces more effective robustness-challenging paraphrases, indicating that the observed gains arise primarily from the proposed training strategy rather than model scale alone. Additional comparisons with larger paraphrase generators show similar trends, with detailed results reported in Appendix~\ref{app:additional-robustness-results}.

These findings suggest that robustness-oriented paraphrase generation requires balancing semantic fidelity and robustness-challenging effectiveness.

\section{Diagnostic Analysis of Robustness Failures}
\label{sec:diagnostic-analysis}

We further analyze whether the robustness failures exposed by \textsc{Hallucination-R1} reflect meaningful factual instability rather than trivial artifacts. Unless otherwise specified, all analyses use semantically consistent paraphrases generated by \textsc{Hallucination-R1} on the SimpleQuestions test set, with Qwen2.5-3B as the QA model.

\subsection{Surface-Level Diagnostics under Semantic Control}

\begin{figure}[t]
    \centering

    \begin{subfigure}[t]{\columnwidth}
        \centering
        \includegraphics[width=\linewidth, trim=0 100 10 100, clip]{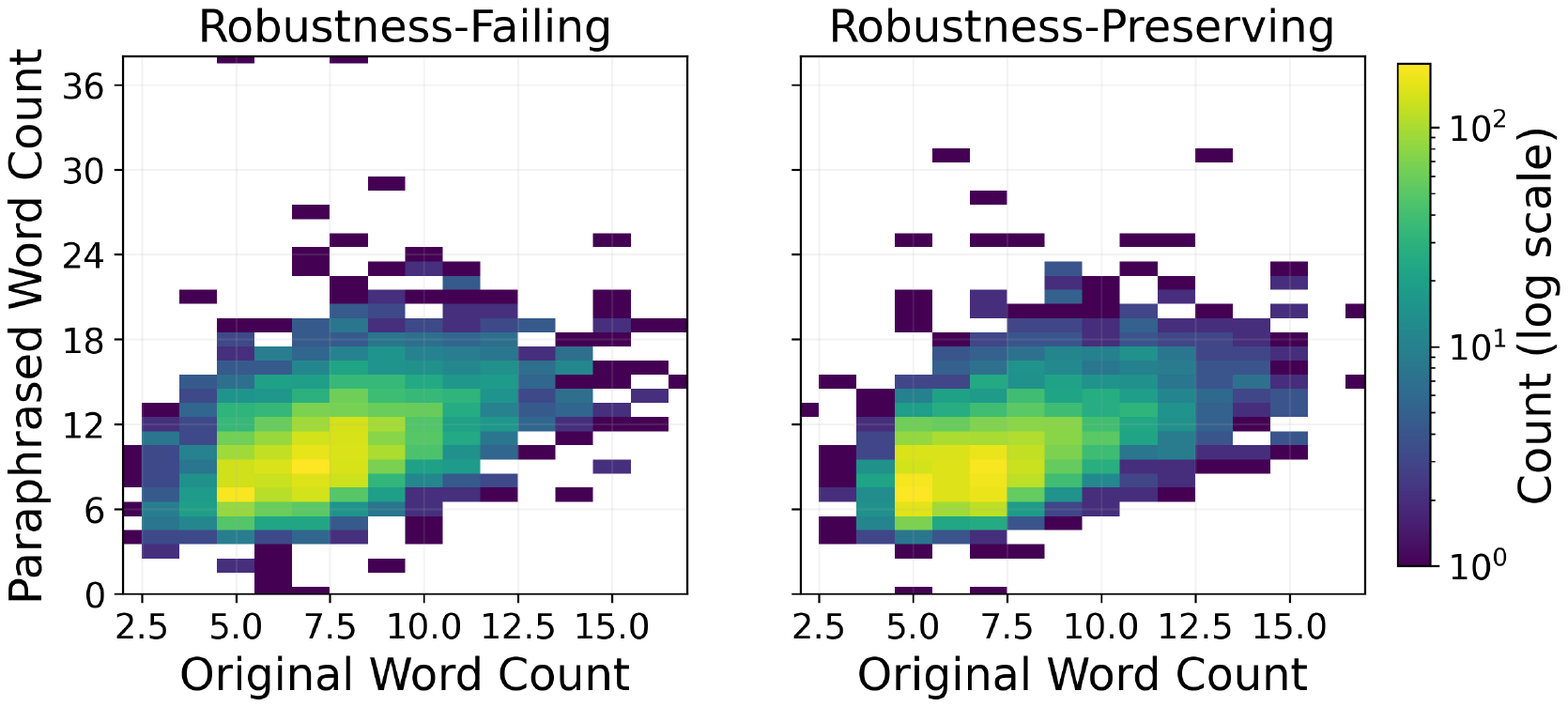}
        \caption{Original vs.\ Paraphrased Question Length}
        \label{fig:hexbin_length}
    \end{subfigure}

    \vspace{6pt}

    \begin{subfigure}[t]{0.40\columnwidth}
        \centering
        \includegraphics[height=2.8cm, trim=80 10 20 10, clip]{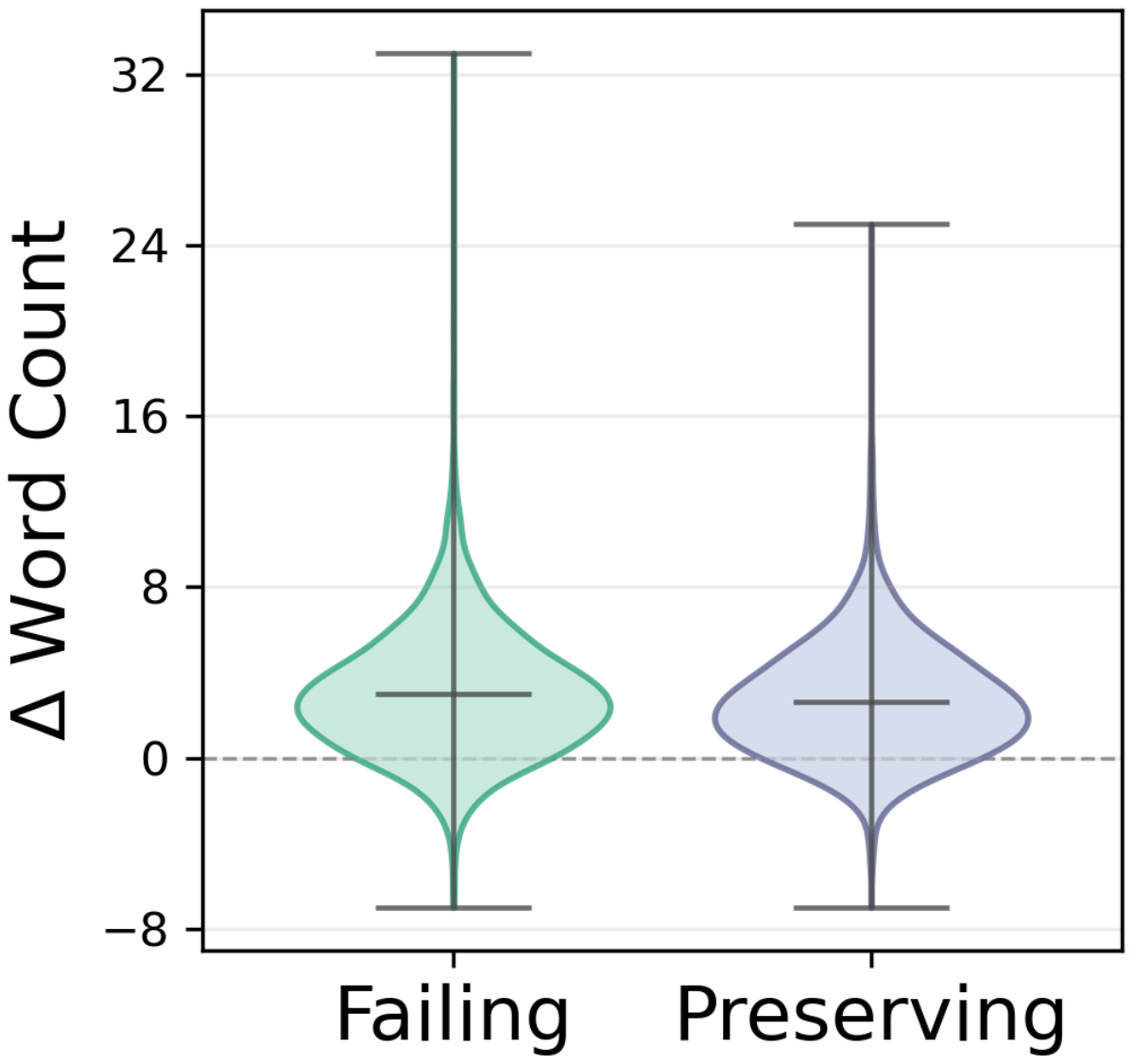}
        \caption{$\Delta$ Word Count}
        \label{fig:delta_word_count}
    \end{subfigure}
    \hfill
    \begin{subfigure}[t]{0.56\columnwidth}
        \centering
        \includegraphics[height=2.8cm, trim=10 10 10 10, clip]{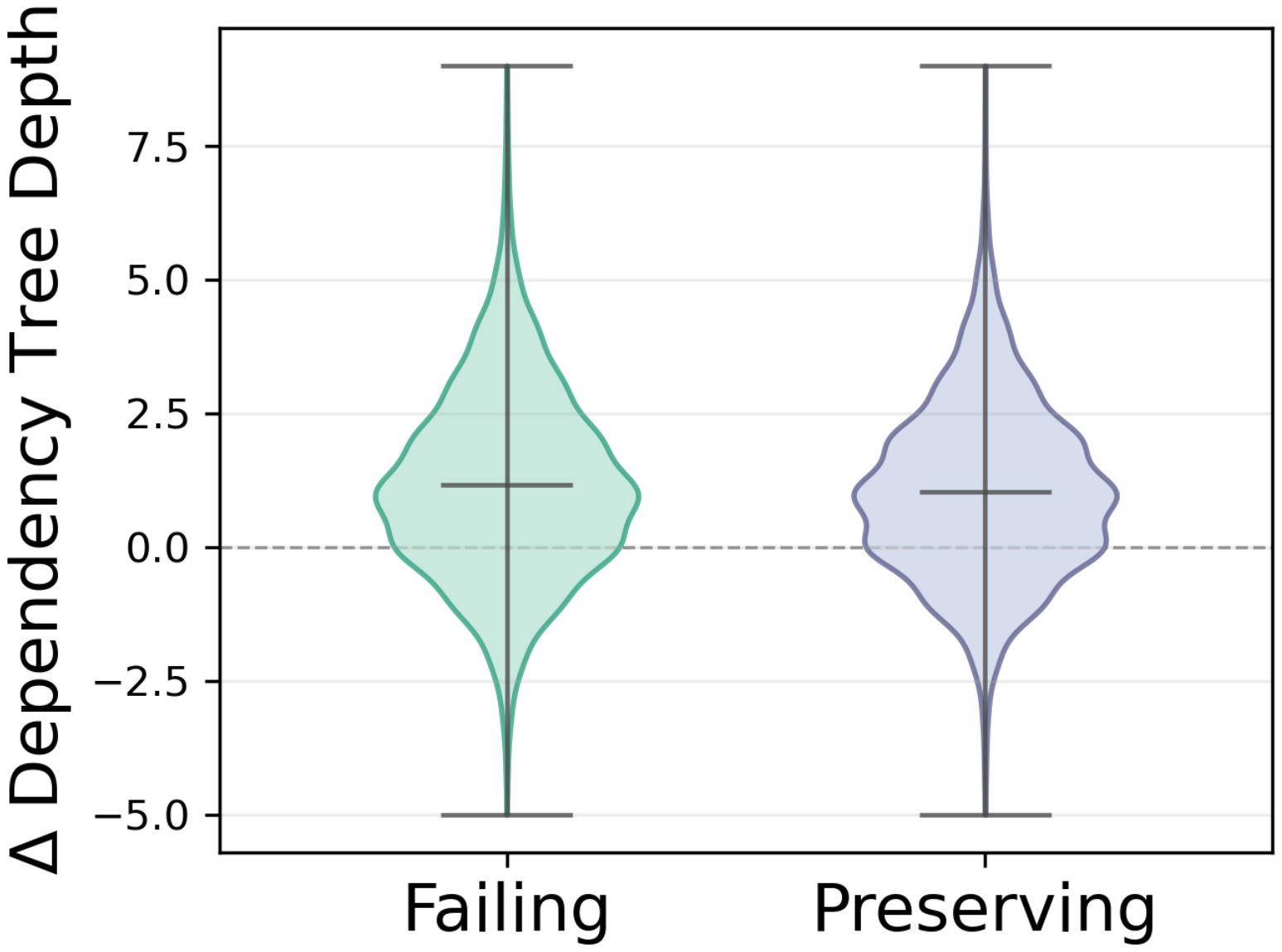}
        \caption{$\Delta$ Dependency Tree Depth}
        \label{fig:delta_dep_depth}
    \end{subfigure}

    \caption{\textbf{Length and structural diagnostics of paraphrases.}
    (a) compares original and paraphrased question lengths for robustness-failing and robustness-preserving cases, denoted as Failing and Preserving in the plots.
    (b) and (c) report changes in word count and dependency tree depth, respectively.
    }
    \label{fig:length_structure}
\end{figure}

\begin{figure}[htpb]
    \centering
    \includegraphics[width=\columnwidth, trim=0 160 0 160, clip]{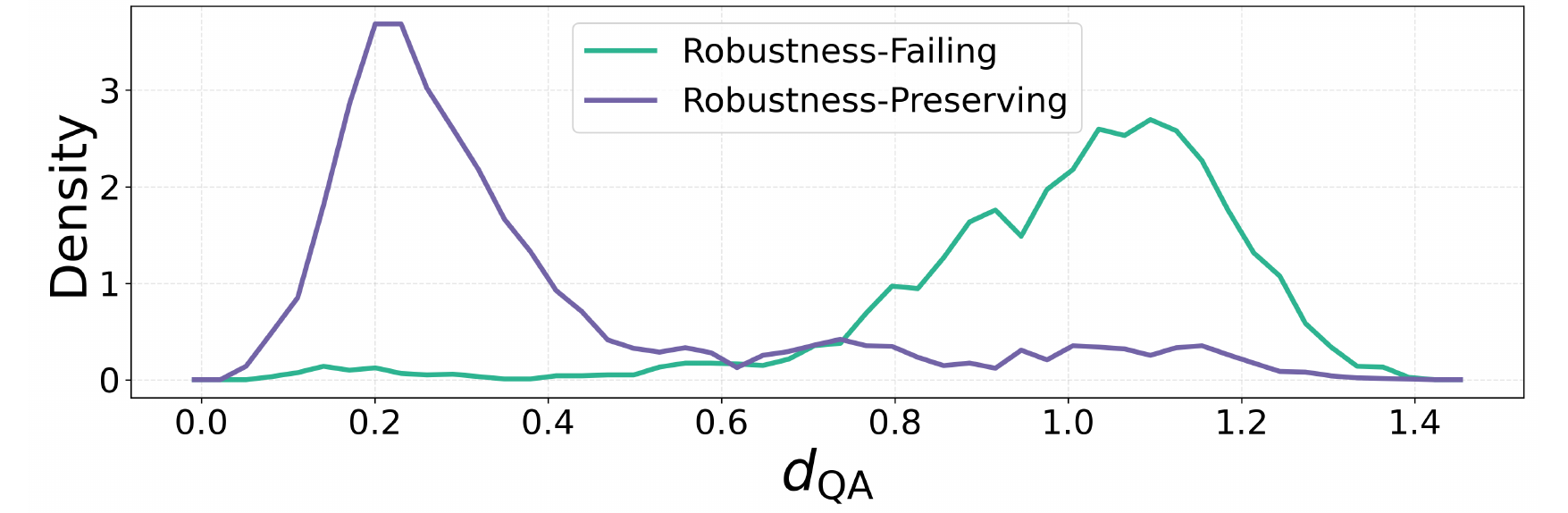}
    \caption{QA-concatenation embedding distance distributions for robustness-failing and robustness-preserving paraphrases.}
    \label{fig:dist_curves}
\end{figure}

We first test whether the observed failures can be explained by simple surface-level changes. As shown in Figure~\ref{fig:length_structure}, robustness-failing and robustness-preserving paraphrases exhibit highly overlapping distributions in word count and dependency tree depth, suggesting that the failures cannot be reduced to longer or syntactically more complex questions. Additional normalized analyses are provided in Appendix~\ref{app:length_and_structure}.

\subsection{Question-Level Similarity versus Response-Level Drift}

We next ask whether question-level semantic similarity is sufficient to ensure response-level factual stability. We compute L2-normalized embedding distances between original and paraphrased questions ($d_Q$) and between their question--answer concatenations ($d_{QA}$); details are given in Appendix~\ref{app:embedding_distance}. As shown in Figure~\ref{fig:dist_curves}, robustness-failing paraphrases are associated with larger $d_{QA}$, whereas preserving paraphrases remain closer to the original QA representations. This suggests that semantic-preserving paraphrases can still induce response-level drift.

This drift is reflected inside the model. In the layer-wise heatmap of Figure~\ref{fig:pipeline}, corresponding to the original and paraphrased questions, we conduct a logit-lens-style analysis and compare the log-probability difference between the first tokens of the correct and incorrect answers across layers. The original question increasingly favors the correct answer in later layers, whereas its semantically equivalent paraphrase shifts toward the incorrect answer. This study shows that the failure is reflected not only in the final response but also in late-stage factual preference dynamics. Additional definitions and examples are provided in Appendix~\ref{app:layer_analysis}.

\begin{table*}[t]
\centering
\resizebox{\linewidth}{!}{
\small
\begin{tabular}{l|l|lll|lll}
\toprule
\multicolumn{1}{c|}{\multirow{2}{*}{\textbf{Paraphrase Model}}} &
\multicolumn{1}{c|}{\multirow{2}{*}{\textbf{QA Model}}}
& \multicolumn{3}{c|}{\textbf{PopQA}}
& \multicolumn{3}{c}{\textbf{SimpleQuestions}} \\
\cmidrule(lr){3-5} \cmidrule(lr){6-8}
& & \textbf{Para. Acc.} 
& \textbf{Robust Acc.} 
& \textbf{Any Acc.}
& \textbf{Para. Acc}. 
& \textbf{Robust Acc}. 
& \textbf{Any Acc.} \\
\midrule
\midrule

\multirow{2}{*}{Hallucination-R1}
& Qwen3-8B 
& 14.30\% & 3.79\% & \textbf{24.32\%} 
& 21.17\% & 6.08\% & \textbf{38.33\%} \\
& \cellcolor{gray!20}Qwen3-8B-SFT  
& \cellcolor{gray!20}\textbf{16.95\%$^{\textcolor{blue}{\uparrow 2.65\%}}$} 
& \cellcolor{gray!20}\textbf{8.06\%$^{\textcolor{blue}{\uparrow 4.27\%}}$} 
& \cellcolor{gray!20}23.03\%$^{\textcolor{red}{\downarrow 1.29\%}}$ 
& \cellcolor{gray!20}\textbf{26.36\%$^{\textcolor{blue}{\uparrow 5.19\%}}$} 
& \cellcolor{gray!20}\textbf{16.93\%$^{\textcolor{blue}{\uparrow 10.85\%}}$} 
& \cellcolor{gray!20}35.44\%$^{\textcolor{red}{\downarrow 2.89\%}}$ \\

\midrule

\multirow{2}{*}{DeepSeek-V3}
& Qwen3-8B 
& 15.90\% & 6.09\% & \textbf{26.47\%} 
& 21.59\% & 5.72\% & \textbf{42.03\%} \\
& \cellcolor{gray!20}Qwen3-8B-SFT  
& \cellcolor{gray!20}\textbf{17.88\%$^{\textcolor{blue}{\uparrow 1.98\%}}$} 
& \cellcolor{gray!20}\textbf{10.62\%$^{\textcolor{blue}{\uparrow 4.53\%}}$} 
& \cellcolor{gray!20}24.49\%$^{\textcolor{red}{\downarrow 1.98\%}}$ 
& \cellcolor{gray!20}\textbf{26.26\%$^{\textcolor{blue}{\uparrow 4.67\%}}$} 
& \cellcolor{gray!20}\textbf{18.33\%$^{\textcolor{blue}{\uparrow 12.61\%}}$} 
& \cellcolor{gray!20}35.96\%$^{\textcolor{red}{\downarrow 6.07\%}}$ \\

\midrule

\multirow{2}{*}{Qwen2.5-32B}
& Qwen3-8B 
& 16.70\% & 7.87\% & \textbf{26.16\%} 
& 22.20\% & 6.33\% & \textbf{41.19\%} \\
& \cellcolor{gray!20}Qwen3-8B-SFT  
& \cellcolor{gray!20}\textbf{18.06\%$^{\textcolor{blue}{\uparrow 1.36\%}}$} 
& \cellcolor{gray!20}\textbf{11.14\%$^{\textcolor{blue}{\uparrow 3.27\%}}$} 
& \cellcolor{gray!20}24.06\%$^{\textcolor{red}{\downarrow 2.10\%}}$ 
& \cellcolor{gray!20}\textbf{26.70\%$^{\textcolor{blue}{\uparrow 4.50\%}}$} 
& \cellcolor{gray!20}\textbf{18.72\%$^{\textcolor{blue}{\uparrow 12.39\%}}$} 
& \cellcolor{gray!20}35.91\%$^{\textcolor{red}{\downarrow 5.28\%}}$ \\

\midrule

\multirow{2}{*}{Qwen3-8B}
& Qwen3-8B 
& 17.09\% & 9.24\% & \textbf{24.52\%} 
& 24.53\% & 10.84\% & \textbf{40.76\%} \\
& \cellcolor{gray!20}Qwen3-8B-SFT  
& \cellcolor{gray!20}\textbf{18.70\%$^{\textcolor{blue}{\uparrow 1.61\%}}$} 
& \cellcolor{gray!20}\textbf{12.87\%$^{\textcolor{blue}{\uparrow 3.63\%}}$} 
& \cellcolor{gray!20}23.03\%$^{\textcolor{red}{\downarrow 1.49\%}}$ 
& \cellcolor{gray!20}\textbf{27.42\%$^{\textcolor{blue}{\uparrow 2.89\%}}$} 
& \cellcolor{gray!20}\textbf{19.11\%$^{\textcolor{blue}{\uparrow 8.27\%}}$} 
& \cellcolor{gray!20}35.90\%$^{\textcolor{red}{\downarrow 4.86\%}}$ \\

\bottomrule
\end{tabular}
}
\caption{
Robustness training results under semantics-preserving paraphrase variations.
Para. Acc., Robust Acc., and Any Acc. are defined in Section~\ref{sec:robustness-training}.
Paraphrase Model denotes the model used to generate evaluation paraphrases, while QA Model denotes the model being evaluated.
$\uparrow$ / $\downarrow$ denote absolute differences in percentage points between Qwen3-8B-SFT and Qwen3-8B under the same paraphrase model and dataset.
}
\label{tab:sft_training}
\end{table*}

\section{Robustness Training Utility}
\label{sec:robustness-training}

A central motivation of our framework is to construct paraphrase data that can serve as effective supervision for robustness-oriented training, rather than merely as probes for robustness evaluation. To examine this training utility, we conduct a lightweight SFT study by fine-tuning Qwen3-8B on semantically consistent and robustness-challenging paraphrases generated by \textsc{Hallucination-R1} from SimpleQuestions. This experiment is intended to assess the utility of the generated data, rather than to propose a complete robustness training framework. Additional implementation details are provided in Appendix~\ref{app:sft-details}.

For evaluation, each original question is paraphrased eight times by each evaluation paraphrase generator, and only semantically consistent paraphrases are retained. The evaluation paraphrase generators include \textsc{Hallucination-R1}, DeepSeek-V3, Qwen2.5-32B, and Qwen3-8B, covering models with strong consistency--diversity trade-offs as well as a high-consistency but conservative paraphrasing baseline. Evaluation is conducted on SimpleQuestions and PopQA.

We report three metrics: Paraphrase Accuracy (Para. Acc.) is computed over individual paraphrased questions; Robust Accuracy (Robust Acc.) counts an original question as correct only if all of its semantically consistent paraphrases are answered correctly; and Any Accuracy (Any Acc.) counts it as correct if at least one paraphrase is answered correctly.

As shown in Table~\ref{tab:sft_training}, fine-tuning with \textsc{Hallucination-R1}-generated data improves Robust Acc., while Para. Acc. changes only modestly and Any Acc. consistently decreases. This pattern is consistent with robustness-oriented training: the model is not primarily encouraged to maximize occasional correctness on at least one variant, but to reduce failures across semantically equivalent variants. In other words, the model becomes better at maintaining factual consistency for questions it is already capable of answering, rather than merely acquiring additional factual knowledge.

Furthermore, the improvement in Robust Acc. generalizes across both datasets and paraphrase models. The gains are not only observed on SimpleQuestions, from which the training data are constructed, but also transfer to PopQA. In addition, similar improvements are observed when evaluation paraphrases are generated by different paraphrase models, including \textsc{Hallucination-R1}, DeepSeek-V3, Qwen2.5-32B, and Qwen3-8B. This indicates that the robustness improvement is not due to adapting to the specific linguistic patterns of \textsc{Hallucination-R1}. Rather, \textsc{Hallucination-R1}-generated data appears to capture broader paraphrase-induced fragility patterns shared across different paraphrasing distributions.

Overall, these results demonstrate the training utility of \textsc{Hallucination-R1}-generated data. Semantically consistent yet challenging paraphrases provide effective robustness-oriented supervision by improving factual consistency under diverse meaning-preserving inputs, rather than primarily expanding the model's factual knowledge.
\section{Conclusion}

We introduced \textsc{Hallucination-R1}, a robustness-oriented paraphrase generation framework for factual consistency under semantic-preserving input variations. By combining semantic stabilization with robustness-challenging optimization, \textsc{Hallucination-R1} generates paraphrases that preserve factual intent while exposing robustness failures in downstream QA models. Experiments across multiple datasets and model families show that the generated paraphrases achieve a strong consistency--diversity trade-off, reveal non-trivial factual instability, and improve robust accuracy when used for lightweight fine-tuning. These results highlight robustness-oriented paraphrase generation as a direction for evaluating and improving factual consistency in large language models.

\section*{Limitations}

This work has three main limitations. First, our evaluation relies on LLM-as-judge for semantic consistency and factual correctness assessment. This reflects a broader limitation in current factual QA and hallucination evaluation, where more authoritative and scalable factual verification signals are often unavailable. Although we conduct additional validation, judge errors or biases may still affect the results.

Second, our robustness-oriented SFT experiment remains preliminary. While the results suggest that semantically consistent yet challenging paraphrases can improve robustness under paraphrase variations, how to systematically use such data to strengthen models against paraphrase-induced factual fragility requires further study.

Third, the paraphrase distribution learned by \textsc{Hallucination-R1} is difficult to explicitly characterize. Our analyses show that the exposed failures cannot be explained by simple surface-level artifacts alone, but what linguistic or representational factors fundamentally make models vulnerable under semantic-preserving variations remains an open question.

\section*{Ethical Considerations}

We use publicly available datasets and model checkpoints for research purposes, following their licenses and terms of use. The generated paraphrases, code, and models released with this work are intended for research on factual robustness evaluation and robustness-oriented training.

\textsc{Hallucination-R1} is designed to study and improve factual robustness under semantics-preserving paraphrase variations. The generated paraphrases are constrained to preserve the original factual query intent and are used in this work for controlled robustness evaluation and robustness-oriented training. The goal is not to create misleading user queries or promote factual errors, but to identify cases where models fail to maintain stable factual behavior under legitimate linguistic variation.

A potential risk is that robustness-challenging paraphrases could be misused to intentionally elicit incorrect answers from deployed models. We mitigate this risk by focusing on factual QA benchmarks with ground-truth answers, reporting aggregate robustness metrics rather than instructions for real-world misuse, and framing the method as a tool for diagnosis and robustness improvement. The paraphrases generated by our framework do not introduce new malicious content or harmful instructions; they reformulate existing factual questions while preserving meaning.

This work also uses automated LLM judges for semantic consistency and factual correctness evaluation. Such judges may introduce biases or errors, so their outputs should not be treated as definitive in high-stakes settings without additional verification. We therefore view \textsc{Hallucination-R1} as a research tool for analyzing and improving factual consistency, rather than as a standalone certification method for model reliability.
\bibliography{ref}

\clearpage
\newpage
\appendix

\section{Methodology Details}
\label{app:methodology-details}

This appendix provides implementation details for the reward computation used in \textsc{Hallucination-R1}. The overall objective and two-stage training pipeline are described in Section~\ref{sec:method}; here we focus on the concrete definitions of semantic consistency, group-relative diversity, and factual inconsistency exposure.

\subsection{Semantic Consistency and Diversity Rewards}
\label{app:warmup-stage}

Given an original question $x$ and a generated paraphrase $y$, we use DeepSeek-V3 to judge whether $y$ preserves the semantic meaning of $x$. The semantic consistency reward is defined as:
\begin{equation}
R_{\text{consis}}(y)=
\begin{cases}
1, & \text{if } M_{\text{consis}}(x, y) = 1, \\
0, & \text{otherwise}.
\end{cases}
\end{equation}
Semantic consistency also serves as a hard gate: semantically inconsistent paraphrases receive zero reward in both training stages.

To encourage diverse paraphrases under this constraint, we adopt a group-relative diversity reward. For each input $x$, we sample a group of paraphrases $\{y_1,\dots,y_G\}$ and retain the subset $Y$ judged semantically consistent. For each $y\in Y$, we compute textual and semantic similarity scores. The textual similarity is based on Self-BLEU:
\begin{equation}
S_{\text{SelfBLEU}}(y)
= -\frac{1}{G}\sum_{n=1}^{G}\text{SelfBLEU}_{Y}(y, n),
\end{equation}
and the semantic similarity is based on cosine similarity between sentence embeddings:
\begin{equation}
S_{\text{embed}}(y)
= -\sum_{y' \in Y}
\frac{\phi(y)\cdot\phi(y')}{\|\phi(y)\|^2\,\|\phi(y')\|^2}.
\end{equation}

We rank candidates by the average similarity score $(S_{\text{SelfBLEU}} + S_{\text{embed}})/2$ in ascending order. The diversity reward is:
\begin{equation}
R_{\text{div}}(y)=
\begin{cases}
\dfrac{\text{Rank}(y, Y)}{|Y|-1}, &
\text{if } R_{\text{consis}}(y) = 1, \\
0, & \text{otherwise}.
\end{cases}
\end{equation}

The semantic stabilization reward follows the definition in Section~\ref{sec:method}:
\begin{equation}
R_{\text{stab}}(y)
=
R_{\text{consis}}(y)
+
\lambda_{\text{div}} R_{\text{div}}(y),
\end{equation}
where $\lambda_{\text{div}}$ is set to $1$ in our experiments.

\subsection{Factual Inconsistency Exposure Reward}
\label{app:harmful-stage}

For each semantically consistent paraphrase $y$, the QA model generates an answer $z \sim \pi_{\text{QA}}(\cdot \mid y)$. 
The LLM-as-judge $M_{\text{halluc}}$ then evaluates the factual inconsistency between $z$ and the ground-truth answer, producing a multi-level inconsistency score:
\begin{equation}
R_{\text{halluc}}(y) = M_{\text{halluc}}(y,z) \in \{0,1,2\}.
\end{equation}
Here, higher values indicate stronger factual inconsistency. 
We use a three-level score rather than a binary correctness label to provide a more graded optimization signal, encouraging the generator to move toward paraphrases that more clearly expose factual robustness failures. 
The detailed judging prompt is provided in Appendix~\ref{app:prompt-judge}.

The final training reward is aligned with Section~\ref{sec:method}:
\begin{equation}
R_{\text{train}}(y)=
\begin{cases}
R_{\text{stab}}(y) + R_{\text{halluc}}(y), & R_{\text{halluc}}\neq0, \\
0, & R_{\text{halluc}}=0.
\end{cases}
\end{equation}

This design preserves semantic consistency as a hard constraint while using $M_{\text{halluc}}$ as a factual inconsistency exposure signal.

\subsection{Training Configuration}
\label{app:training-config}

\paragraph{Model Setup}
We consider two distinct model roles during training: the QA model whose factual robustness is evaluated and the \textsc{Hallucination-R1} paraphrasing model. These two models are configured with different decoding
strategies to reflect their functional objectives.

For the QA model, we disable stochastic decoding and adopt deterministic generation throughout all
training and evaluation phases (i.e., \texttt{do\_sample=False}). This design ensures that the QA model’s responses are stable and reproducible given a fixed input, eliminating randomness as a confounding factor when attributing factual errors to paraphrased queries. As a result, any observed factual failure can be reliably attributed to the meaning-preserving paraphrase rather than decoding noise.

In contrast, the \textsc{Hallucination-R1} model is trained and deployed with stochastic sampling enabled to
encourage exploration of diverse paraphrasing strategies. Specifically, we adopt a sampling-based
decoding configuration with \texttt{do\_sample=True}, temperature-controlled generation, and nucleus
sampling. The generation function is defined as:
\begin{listingbox}[label={Hallucination-R1-arg}]{Hallucination-R1 configuration}
\[
\begin{aligned}
\texttt{do\_sample}   &= \texttt{True} \\
\texttt{temperature} &= 0.7 \\
\texttt{max\_tokens} &= 1024 \\
\texttt{top\_k}      &= 0 \\
\texttt{top\_p}      &= 0.95
\end{aligned}
\]
\end{listingbox}

This asymmetric decoding design is crucial: stochastic paraphrase generation enables broad exploration, while deterministic QA decoding ensures that observed factual errors arise from meaning-preserving input variations rather than QA output randomness.

\subsection{Training Dynamics}
\label{app:training-dynamics}

\begin{figure}[htpb]
    \centering
    \includegraphics[width=\linewidth]{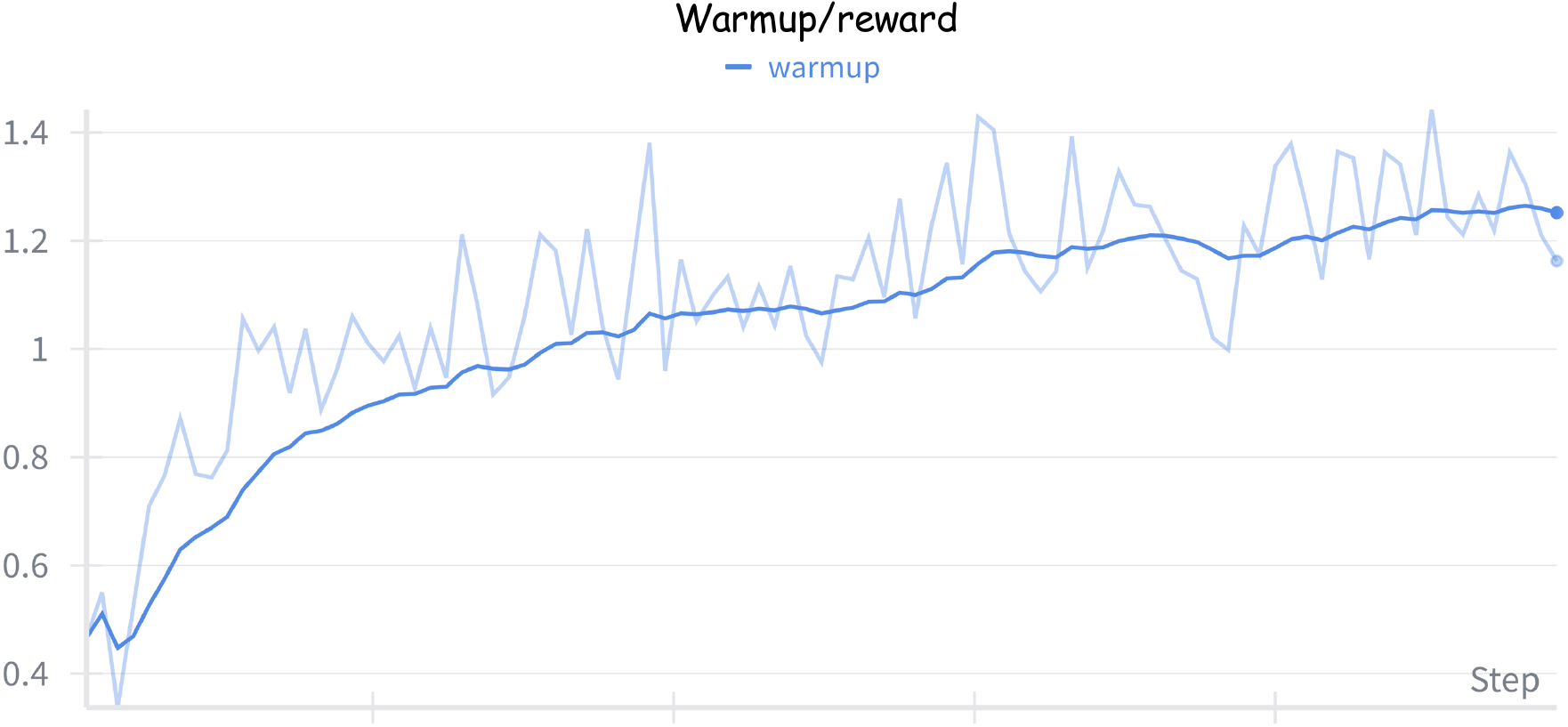}
    \caption{Training reward curve during the semantic stabilization stage.}
    \label{fig:warmup-reward}
\end{figure}

\paragraph{Semantic Stabilization Training Results}
Figure~\ref{fig:warmup-reward} shows the reward curve during semantic stabilization. We train for one epoch on 1,536 samples, jointly optimizing semantic consistency and surface-level diversity. Although individual update steps exhibit variance due to GRPO sampling, the smoothed reward steadily increases from approximately 0.4 to above 1.2.

This trend indicates that the model gradually learns to preserve the original query meaning while exploring alternative syntactic and stylistic forms. It also provides a well-conditioned initialization for the subsequent robustness-challenging optimization stage.

\begin{figure}[htpb]
    \centering
    \includegraphics[width=\linewidth]{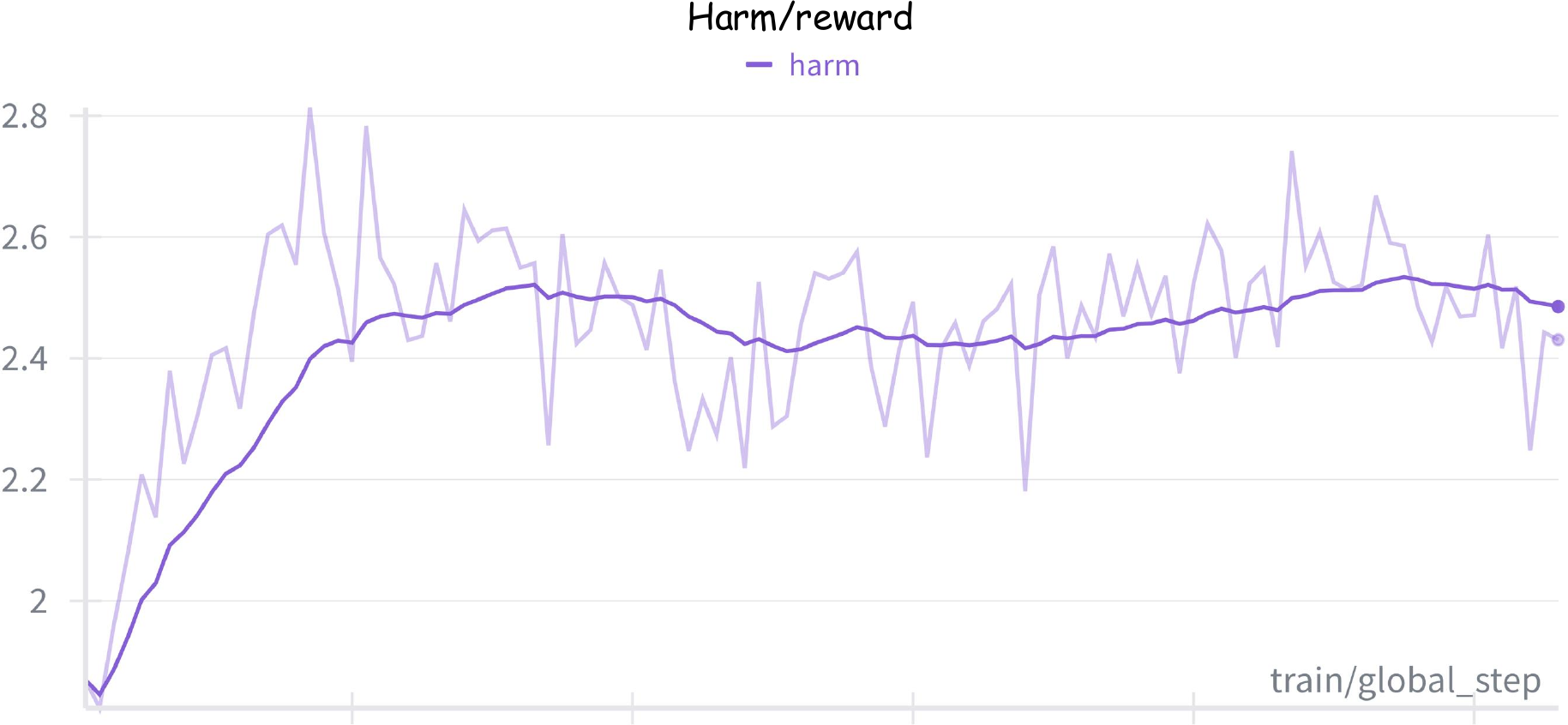}
    \caption{Training reward curve during the robustness-challenging optimization stage.}
    \label{fig:harm-reward}
\end{figure}

\paragraph{Robustness-Challenging Optimization Dynamics}
After semantic stabilization, we conduct robustness-challenging optimization to maximize failure exposure under semantic invariance. As shown in Figure~\ref{fig:harm-reward}, the reward increases rapidly at the beginning and then enters a high-variance plateau, suggesting that the model first discovers basic failure-exposing patterns and then relies on finer-grained interactions with the QA model's decision boundaries.

Despite step-level variance, the smoothed reward remains stable around 2.5. The slight downward trend near the end suggests a trade-off between exploring more complex linguistic forms and maintaining semantic-preservation constraints. Overall, the second stage improves failure exposure while retaining controllability.

\paragraph{Training Infrastructure}
All training experiments were conducted on a single server equipped with 8 NVIDIA A800 80GB GPUs. 
We trained two separate \textsc{Hallucination-R1} models following the same training pipeline. 
For each model, the semantic stabilization stage required approximately 8 hours, and the robustness-challenging optimization stage required approximately 20 hours under the default configuration described above.

\section{Prompt Templates}
\label{app:prompt}

\subsection{Paraphrase Prompt}
\label{app:prompt-attack}

\begin{listingbox}[label={Robustness-Paraphrase-Prompt}]{Paraphrase Prompt}
Your task is to paraphrase a factual question to expose weaknesses of the QA model under semantic invariance, while STRICTLY preserving the exact semantic meaning and intent of the original question.

CRITICAL CONSTRAINTS:\\
- Do NOT add any extra questions, conditions, or information.\\
- Do NOT change any key factual elements (tasks, entities, time, location, numbers,
  relationships, etc.).\\
- Do NOT change what is being asked --- the answer to the paraphrased question must remain
  identical to the original.\\
- ONLY modify: word order, sentence structure, phrasing style, grammatical construction,
  linguistic presentation.\\
Your output should include:\\
1. Verify the core semantic elements that MUST be preserved (what is being asked, about
   what/whom, under what conditions).\\
2. Identify stylistic transformation techniques that could make the question more challenging for factual QA WITHOUT
   changing meaning (e.g., passive voice, complex syntax, double negatives, nested clauses,
   unusual word order, formal/archaic language, indirect phrasing, pronoun substitution,
   nominalization).\\
3. Generate the paraphrased question that maintains semantic consistency but uses confusing
   stylistic patterns.\\
Output format:\\
\texttt{<think>} your analysis and paraphrasing strategy goes here \texttt{</think>}\\
\texttt{<paraphrase>} ONLY the paraphrased question, no explanations or analysis \texttt{</paraphrase>}\\
Original question: \{question\}
\end{listingbox}

\subsection{Judge Prompt}
\label{app:prompt-judge}

\begin{listingbox}[label={Consistency-Prompt}]{Consistency Judge Prompt}
Please help me determine whether the paraphrased question generated by the paraphrase model has the same semantic meaning as the original question.

CRITERIA FOR SEMANTIC CONSISTENCY:\\
- The paraphrased question must NOT add any extra questions, conditions, or information beyond the original. \\
- The paraphrased question must NOT change any key factual elements (tasks, entities, time, location, numbers, relationships, etc.). \\
- The primary entities and query intent of the original question must be explicitly included in the paraphrased question (substituting similar words is acceptable). \\
- Modifying phrasing style, sentence structure, or word order is acceptable as long as the core question remains the same. \\
- The paraphrased question must NOT contain excessively long analysis, explanations, or irrelevant text beyond the question itself. \\

EVALUATION PROCESS:\\
1. Extract the core factual query that the paraphrased question is asking.\\
2. Check if any extra questions, conditions, or information have been added — if YES, result is 0. \\
3. Check if any key factual elements have been changed — if YES, result is 0. \\
4. Determine whether the original question's semantic intent is preserved — if YES,  result is 1; otherwise result is 0. \\

Output format: {"think": your process of thinking, "result": 1 or 0}

Original question: {question}

paraphrased question: {paraphrase}

Output Only: {"think": your process of thinking, "result": 1 or 0}
\end{listingbox}







\begin{listingbox}[label={Correctness-Prompt}]{Factual Inconsistency Judge Prompt}
Your task is to evaluate whether the model's response correctly answers the question,
based on the provided reference answer.

Use a 3-level score:

Score 0 (Exact Match): The response matches the reference answer exactly, or is
semantically equivalent to the same entity.

Score 1 (Fuzzy Match): The response and the reference answer have a parent-child
class relationship, or one is an abbreviated form of the other.

Score 2 (Mismatch): The response contradicts the reference answer, is factually
wrong, or refers to a completely different entity.

This is part of an automated evaluation process. Therefore, the output must be a
single JSON object in the following format:
\{"think": "your reasoning process", "result": 0/1/2\}

Reference Answer: \{ground\_truth\}

Model Response: \{response\}

Output only:
\{"think": "your reasoning process", "result": 0/1/2\}
\end{listingbox}








\subsection{QA Model Answer Prompt}
\label{app:prompt-target}

\begin{listingbox}[label={Target-Answer-Prompt}before skip=-6pt]{QA Model Answer Prompt}
You are a strict question answering model. \\
You MUST output ONLY the final answer inside the \texttt{<answer>} and \texttt{</answer>} tags. \\
Do NOT repeat the question.\\
Do NOT explain. \\
Do NOT output anything outside these tags. 
\end{listingbox}


\section{Experimental Setup}
\label{app:experimental-setup}

\subsection{Paraphrase Generator Training Data}
\label{app:paraphrase-training-data}

The training data for \textsc{Hallucination-R1} are derived from the training split of SimpleQuestions. To ensure that factual errors observed during training can be attributed to meaning-preserving paraphrase variations, we first run Qwen2.5-3B as the initial QA model ($\pi_{\text{QA}}$) on the training set and retain only instances that it answers correctly. After filtering, approximately 7{,}000 question--answer pairs remain for training the paraphrase generator $\pi_{\text{para}}$.

\subsection{Paraphrase Generator Training Setup}
\label{app:paraphrase-training-setup}

We train two paraphrase generators based on Qwen2.5-1.5B and Qwen2.5-3B, respectively. Both models are optimized with the two-stage pipeline described in Section~\ref{sec:method}, yielding \textsc{Hallucination-R1-1.5B} and \textsc{Hallucination-R1-3B}. 

Throughout the paper, we use \textsc{Hallucination-R1} to denote \textsc{Hallucination-R1-3B} unless a specific model scale is explicitly stated.

\subsection{Evaluation Datasets}
\label{app:evaluation-datasets}

We evaluate on SimpleQuestions, PopQA, and TruthfulQA. Table~\ref{tab:datasets} summarizes the dataset sizes and evaluation protocols.

\begin{table}[htpb]
\centering
\resizebox{\linewidth}{!}{
\begin{tabular}{lccc}
\hline
\textbf{Dataset} & \textbf{Total Samples} & \textbf{Evaluation Metrics} \\
\hline
SimpleQuestions & 9,910  & LLM Judge \\
PopQA  & 14,267 & EM Match \\
TruthfulQA     & 790    & LLM Judge \\
\hline
\end{tabular}
}
\caption{Evaluation datasets and metrics.}
\label{tab:datasets}
\end{table}

\subsection{Robustness Evaluation Protocol}
\label{app:robustness-evaluation-protocol}

Algorithm~\ref{alg:robustness_eval} summarizes the robustness evaluation procedure used to compute ER@1 and WER@5. A robustness failure is counted only when the paraphrased question is judged semantically equivalent to the original question and the QA model produces an incorrect answer.

\begin{algorithm}[t]
\caption{Robustness Evaluation under Semantic-Preserving Paraphrases}
\label{alg:robustness_eval}
\begin{algorithmic}[1]
\Require Question $x$, ground-truth answer $y$
\Ensure Number of robustness failures $\text{ERR}$

\State $\text{ERR} \gets 0$
\State $x' \gets \pi_{\text{para}}(x)$

\If{$M_{\text{consis}}(x, x') = True$}
    \State $z \gets \pi_{\text{tgt}}(x')$
    \If{$M_{\text{judge}}(x, y, z) = False$}
        \State $\text{ERR} \gets \text{ERR} + 1$
    \EndIf
\EndIf

\State $\text{ErrorRate} \gets \text{ERR} / \text{Num. of True Samples}$

\State \Return $\text{ERR}$, $\text{ErrorRate}$

\end{algorithmic}
\end{algorithm}

\subsection{Compared Paraphrase Models}
\label{app:compared_models}

In the consistency--diversity analysis, we compare \textsc{Hallucination-R1} with both open-source paraphrase generators and prompted LLM paraphrasing baselines.
For trainable paraphrase generators, we include the Qwen2.5-1.5B and Qwen2.5-3B backbones, as well as their intermediate variants obtained after the Semantic Stabilization Stage.
These models are used to examine how our two-stage training pipeline changes the paraphrase distribution relative to the base generators.

For prompted baselines, we compare against stronger instruction-tuned LLMs, including Qwen2.5-7B, Qwen2.5-32B \cite{qwen2025qwen25technicalreport}, Qwen3-4B \cite{yang2025qwen3}, LLaMA-3.1-8B \cite{grattafiori2024llama}, DeepSeek-V3, DeepSeek-R1 \cite{guo2025deepseek}, and GPT-5 \cite{singh2025openai}.
All prompted baselines are evaluated using the same paraphrase-generation instruction and the same consistency/diversity metrics as \textsc{Hallucination-R1}.

Unless otherwise specified, the QA models used in evaluation follow the same inference configuration as the QA model used for training feedback.
We disable sampling for QA inference to ensure that observed factual errors arise from input-level paraphrase variation rather than output randomness.
In contrast, paraphrase generation uses stochastic decoding to explore diverse paraphrases.

\section{Additional Robustness Evaluation Results}
\label{app:additional-robustness-results}

\subsection{Effect of Model Scale and Training Strategy}
\label{app:scale-training-comparison}

\begin{table*}[htpb]
\centering
\resizebox{\linewidth}{!}{
\begin{tabular}{l c | l l c c l l}
\toprule
\textbf{QA Model}
& \textbf{Ori. Acc}
& \textbf{Paraphrase Model}
& \textbf{Consis@1}
& \textbf{RF@1}
& \textbf{RF@5}
& \textbf{ER@1}
& \textbf{WER@5} \\
\midrule
\multirow{6}{*}{Qwen2.5-7B}
& \multirow{6}{*}{323 (40.89\%)}
& Qwen2.5-3B
& 64.68\% & 98 & 233 & 30.34\% & 72.14\% \\

&
& Qwen2.5-7B
& 77.32\% & 124 & 262 & 38.39\% & 81.11\% \\

&
& Qwen2.5-32B
& 93.17\% & 135 & 256 & 41.80\% & 79.26\% \\

&
& DeepSeek-V3
& 97.72\% & 164 & 273 & 50.77\% & 84.52\% \\

&
& Hallucination-R1 (Stage 1 Only)
& 80.37\% & 125 & 245 & 38.70\% & 75.85\% \\

&
& \cellcolor{gray!20}\textbf{Hallucination-R1-3B}
& \cellcolor{gray!20}\textbf{87.46\%}
& \cellcolor{gray!20}\textbf{147}
& \cellcolor{gray!20}\textbf{277}
& \cellcolor{gray!20}\textbf{45.51\%}
& \cellcolor{gray!20}\textbf{85.76\%} \\
\bottomrule
\end{tabular}
}
\caption{Additional results on TruthfulQA with Qwen2.5-7B as the QA model. We compare \textsc{Hallucination-R1-3B} with paraphrase models of different scales and architectures.}
\label{tab:additional-qwen7b}
\end{table*}


To examine whether the ability to expose factual inconsistency comes primarily from model scale or from the proposed training strategy, we conduct additional comparisons on TruthfulQA using Qwen2.5-7B as the QA model. As shown in Table~\ref{tab:additional-qwen7b}, \textsc{Hallucination-R1-3B} achieves stronger ER@1 and WER@5 than larger untrained paraphrase models, including Qwen2.5-7B and Qwen2.5-32B, while maintaining high semantic consistency. Compared with the Stage-1-only variant, the full model also yields substantially higher failure exposure, showing the contribution of the robustness-challenging optimization stage.
These results suggest that the effectiveness of \textsc{Hallucination-R1} stems from the proposed training strategy rather than parameter scale alone.

\section{Robustness-Oriented SFT Details}
\label{app:sft-details}

This appendix provides implementation details for the robustness-oriented SFT experiment in Section~\ref{sec:robustness-training}.

\subsection{SFT Data Construction}
\label{app:sft-data}

The SFT data are constructed directly from the training split of SimpleQuestions, without filtering original questions based on QA model correctness. For each original question, \textsc{Hallucination-R1} generates multiple robustness-challenging paraphrases. We retain only paraphrases that pass the semantic consistency verifier and pair each retained paraphrase with the original ground-truth answer. This construction provides robustness-oriented supervision over meaning-preserving surface variations while relying only on the original dataset annotations.

\subsection{SFT Training Setup}
\label{app:sft-training}

We fine-tune Qwen3-8B with LoRA-based supervised fine-tuning. The objective is the standard next-token prediction loss over the ground-truth answer, conditioned on robustness-challenging paraphrased questions. Only LoRA parameters are updated, while the base model remains frozen.

\subsection{SFT Evaluation Protocol}
\label{app:sft-eval}

For evaluation, we follow Section~\ref{sec:robustness-training}: each original question is paraphrased eight times by each paraphrase model, and only semantically consistent paraphrases are retained. We report Paraphrase Acc., Any Acc., and Robust Acc. Robust Acc. is computed at the original-question level and requires all retained paraphrases of the same original question to be answered correctly.

\section{Judge Consistency Validation}

\subsection{Cross-Model Consistency Validation}
\label{app:cross-consistency}

To verify that semantic consistency evaluation does not depend on a single judge model, we conduct a cross-model validation experiment with LLM judges of different architectures. Each judge independently evaluates the paraphrased samples and produces binary consistency decisions. We report \textbf{Consis@1}, \textbf{Consis@5}, and \textbf{Majority@5}, following the definitions in the main paper.

As shown in Table~\ref{tab:cross-consistency-results}, the consistency judgments are highly aligned across judge models. This suggests that the semantic consistency evaluation is robust to the choice of judge model and reflects shared agreement across different LLMs.

\subsection{Human Consistency Annotation Study}
\label{app:human-consistency}

To further rule out shared biases among LLM judges, we conduct a small-scale human annotation study. We randomly sample 100 original questions from the TruthfulQA paraphrased set, yielding 500 question--paraphrase pairs. Each pair is annotated by at least three graduate students with NLP backgrounds recruited for a small-scale expert validation study. No monetary compensation was provided, and only aggregate annotation results are reported.

Annotators only see the original and paraphrased questions. They were instructed: ``Given an original question and a paraphrased question, decide whether the paraphrase preserves the original querying intent and factual reference. Mark the pair as consistent only if it does not add, remove, or change key entities, relations, conditions, or the answer-bearing intent.'' The final label is determined by majority voting.

As shown in Table~\ref{tab:cross-consistency-results}, human judgments are broadly aligned with LLM-based consistency judgments. This provides additional evidence that the semantic consistency evaluation is not solely driven by shared LLM judge biases, but is also consistent with human intuition.

\begin{table}[htpb]
\centering
\resizebox{\linewidth}{!}{
\begin{tabular}{lccc}
\hline
\textbf{Judge Model} & \textbf{Consis@1} & \textbf{Consis@5} & \textbf{Majority@5} \\
\hline
DeepSeek-V3~\cite{liu2024deepseek}     & 87.46\% & 99.87\% & 97.09\% \\
Kimi-K2~\cite{guo2025deepseek}         & 83.72\% & 99.49\% & 95.87\% \\
Qwen3-235B-A22B~\cite{yang2025qwen3} & 85.82\% & 100.00\% & 96.46\% \\
\rowcolor{gray!15}
Human (100 samples) & 83.00\% & 98.00\% & 96.00\% \\
\hline
\end{tabular}
}
\caption{Semantic consistency evaluation on the TruthfulQA dataset across different judge models and human annotation.}
\label{tab:cross-consistency-results}
\end{table}

\section{Supplementary Analysis}
\label{app:supplementary_analysis}

This section provides supplementary analyses for
Section~\ref{sec:diagnostic-analysis}. We expand the surface-level diagnostics,
embedding-distance analysis, and layer-wise factual preference analysis used to
examine whether robustness failures exposed by \textsc{Hallucination-R1} can be
attributed to trivial artifacts or reflect non-trivial factual instability.

\subsection{Length and Structure Features}
\label{app:length_and_structure}

\begin{figure}[htpb]
    \centering
    \includegraphics[width=\columnwidth, trim=60 20 60 0, clip]{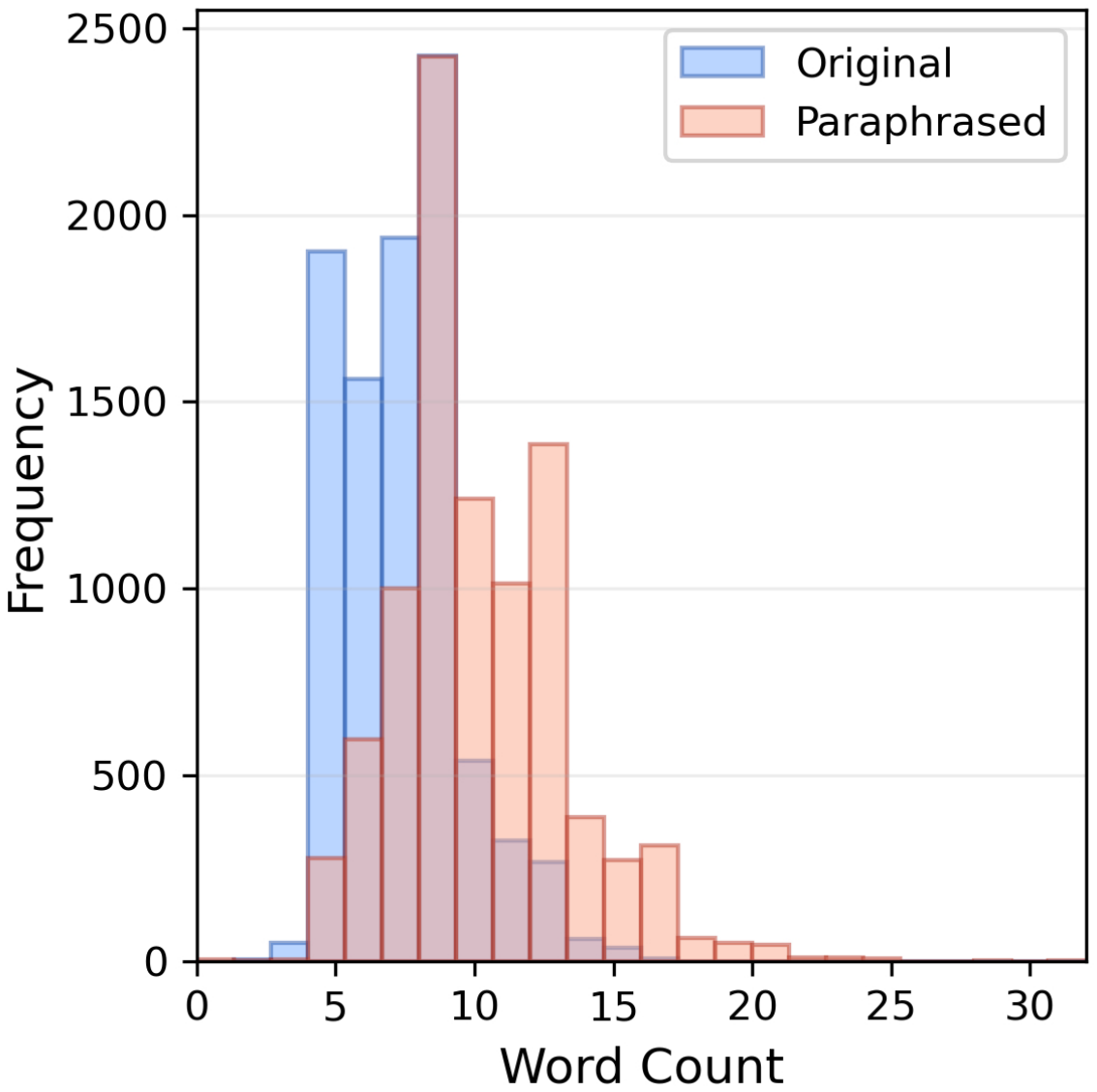}
    \caption{Distribution of word counts for original and paraphrased questions across all questions.}
    \label{fig:overlay_hist}
\end{figure}

\paragraph{Sentence Length Distributions.}
As shown in Figure~\ref{fig:overlay_hist}, \textsc{Hallucination-R1} exhibits a clear tendency to increase sentence length. However, sentence length alone does not clearly distinguish robustness-failing from robustness-preserving paraphrases.

\paragraph{Dependency Tree Depth.}
Dependency tree depth in Figure~\ref{fig:delta_dep_depth} is computed using an off-the-shelf parser based on the Universal Dependencies (UD) formalism. 
We define tree depth as the maximum distance from the root token to any node in the dependency tree. 
Our analysis focuses exclusively on relative depth changes ($\Delta$) induced by paraphrasing, rather than absolute depth values.

\begin{figure}[htpb]
    \centering
    \includegraphics[width=\columnwidth, trim=0 70 0 0, clip]{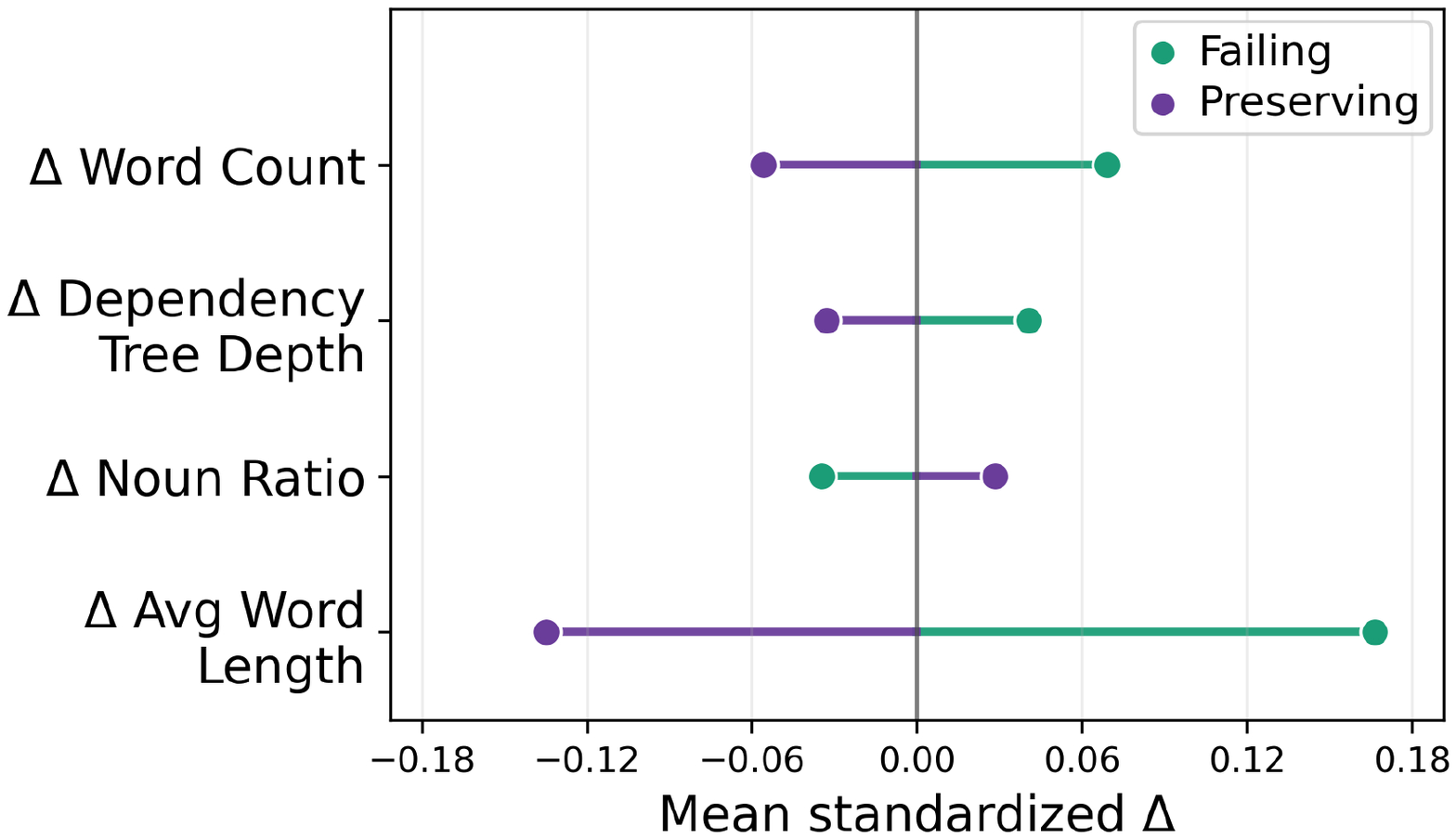}
    \caption{Normalized feature profiles of robustness-failing and robustness-preserving paraphrases under the global distribution.}
    \label{fig:horizontal_profile}
\end{figure}

\paragraph{Normalized Structural Feature Profiles.}
Figure~\ref{fig:horizontal_profile} presents normalized feature profiles for robustness-failing and robustness-preserving paraphrases under a global distributional reference.

We refer to paraphrases that lead the QA model to incorrect answers as robustness-failing (RF) paraphrases, and those that preserve correct QA behavior as robustness-preserving (RP) paraphrases.

\emph{Feature definitions.}
Surface length and lexical form complexity are treated as complementary signals.
Changes in sentence length are measured by the word-count difference:
\begin{equation}
\Delta \textsc{WC}
= |\mathcal{W}^{\text{para}}| - |\mathcal{W}^{\text{orig}}|,
\label{eq:delta_wc}
\end{equation}
where $|\mathcal{W}|$ denotes the number of words in a question.
In contrast, changes in average word length capture shifts in lexical morphology (e.g., preferences for longer or more complex word forms):
\begin{equation}
\Delta \textsc{AWL}
= \frac{|\mathcal{C}^{\text{para}}|}{|\mathcal{W}^{\text{para}}|}
- \frac{|\mathcal{C}^{\text{orig}}|}{|\mathcal{W}^{\text{orig}}|},
\label{eq:delta_awl}
\end{equation}
where $|\mathcal{C}|$ denotes the number of characters.
While $\Delta \textsc{WC}$ reflects the external scale of sentence length (information volume), $\Delta \textsc{AWL}$ isolates changes in word-form complexity independent of sentence length.

\emph{Normalization procedure.}
For each feature dimension, standardized values are computed with respect to the overall distribution across all paraphrases.
Let $\Delta f_i$ denote the change value of a structural feature (e.g., word count, dependency tree depth, noun ratio) for sample $i$.
The global mean and standard deviation are defined as:
\begin{equation}
\mu_f = \frac{1}{N} \sum_{i=1}^{N} \Delta f_i,
\end{equation}
\begin{equation}
\sigma_f = \sqrt{ \frac{1}{N} \sum_{i=1}^{N} (\Delta f_i - \mu_f)^2 },
\end{equation}
where $N$ denotes the total number of labeled samples.

Each sample is then mapped to a standardized score:
\begin{equation}
z_i = \frac{\Delta f_i - \mu_f}{\sigma_f}.
\end{equation}

For visualization, mean standardized scores are reported separately for robustness-failing and robustness-preserving paraphrases:
\begin{equation}
\bar{z}_{\text{RF}}=\mathbb{E}[z_i \mid i \in \text{RF}],
\quad
\bar{z}_{\text{RP}}=\mathbb{E}[z_i \mid i \in \text{RP}].
\end{equation}

Although absolute feature distributions largely overlap, normalization reveals consistent directional shifts in group-wise means.
In particular, robustness-failing paraphrases exhibit small but systematic tendencies across multiple dimensions, including slightly increased sentence length, moderately higher syntactic complexity, and a reduced inclination toward nominalization.

\subsection{Embedding Distance}
\label{app:embedding_distance}

We detail the embedding distance computation.
Both original and paraphrased questions, as well as their corresponding generated answers, are formatted using the following templates: 

\begin{listingbox}[label={lst:embedding_format}]{Embedding Input Templates}
\begin{center}
\begin{minipage}{0.95\columnwidth}
\small
\begin{tabular}{@{}l l@{}}
\textit{Question:} &
\texttt{\detokenize{"Question: {question}"}}\\[0.5em]

\textit{QA-concatenation:} &
\texttt{\detokenize{"Question: {question}\n"}}\\
& \texttt{\detokenize{"Answer: {answer}"}}
\end{tabular}
\end{minipage}
\end{center}
\end{listingbox}

All formatted texts are fed into Qwen2.5-3B.
Hidden states are extracted from the 18th transformer layer (out of 36 total layers), and the hidden state of the final token is used as the sequence embedding.
Pairwise Euclidean (L2) distances between embeddings are computed and normalized.

Based on the above distance definitions, we further define a relative distance difference:
\begin{equation}
\Delta = d_{QA} - d_Q.
\end{equation}
A positive $\Delta$ indicates that the generated response amplifies the semantic deviation introduced by paraphrasing, whereas a negative $\Delta$ suggests that the response pulls the joint QA representation back toward the original semantic neighborhood.

\begin{figure}[t]
    \centering
    \begin{subfigure}{0.9\linewidth}
        \centering
        \includegraphics[width=\linewidth, trim=50 20 20 10, clip]{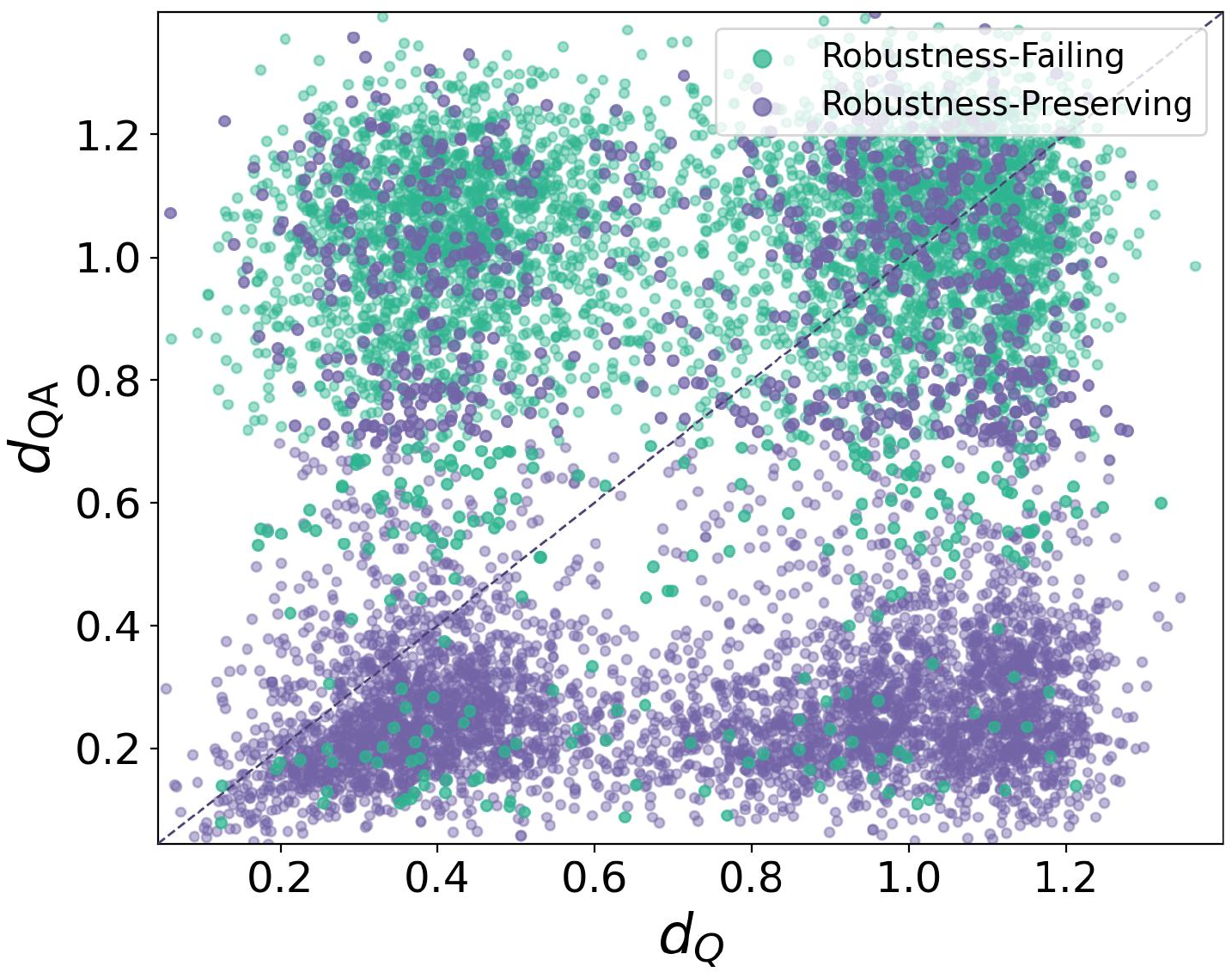}
        \caption{Joint distribution of $d_Q$ and $d_{QA}$}
        \label{fig:scatter_delta_a}
    \end{subfigure}

    \begin{subfigure}{0.9\linewidth}
        \centering
        \includegraphics[width=\linewidth, trim=45 20 0 0, clip]{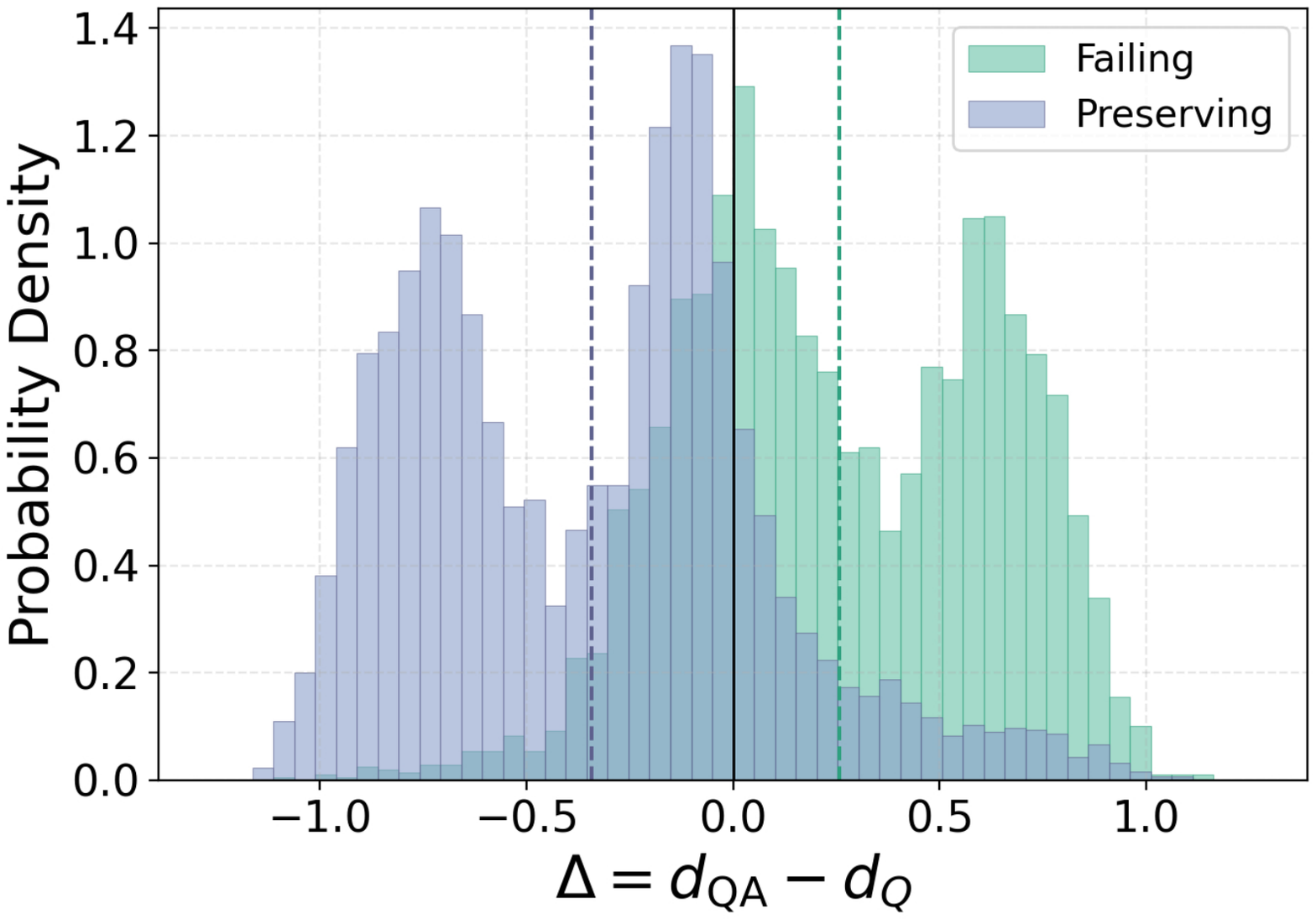}
        \caption{Distribution of $\Delta$}
        \label{fig:scatter_delta_b}
    \end{subfigure}

    \caption{\textbf{Relationship between question paraphrase distance and QA-concatenation distance.}
    Robustness-failing and robustness-preserving paraphrases exhibit distinct patterns in both the joint distance space and the induced $\Delta$ distribution.}
    \label{fig:scatter_delta}
\end{figure}

Under this definition, robustness-failing paraphrases predominantly lie in regions with positive $\Delta$, while robustness-preserving paraphrases cluster around negative or near-zero $\Delta$, a pattern further confirmed by the $\Delta$ histogram (Figure~\ref{fig:scatter_delta}).

\subsection{Layer-wise Factual Preference Dynamics}
\label{app:layer_analysis}

\begin{figure*}[t]
    \centering
    \includegraphics[width=\linewidth, trim=20 0 10 0, clip]{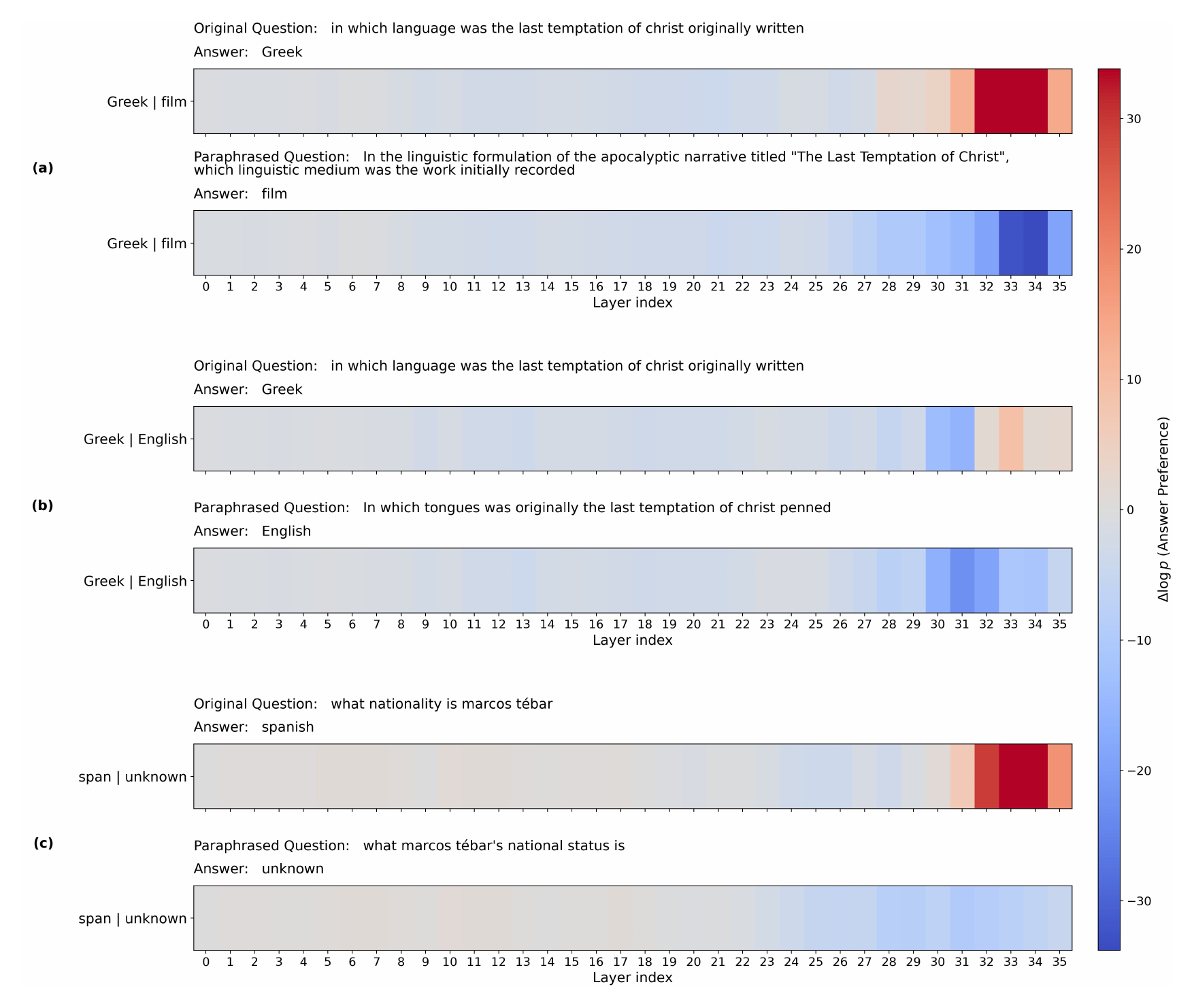}
    \caption{Additional examples of layer-wise factual preference dynamics under original and paraphrased inputs.}
    \label{fig:layer_heatmap_3}
\end{figure*}

We provide additional details for the layer-wise case study shown in Figure~\ref{fig:pipeline} and present further examples in Figure~\ref{fig:layer_heatmap_3}. We conduct a logit-lens-style analysis by projecting the hidden state at each layer to the vocabulary space using the model's output head. This allows us to inspect how the model's preference between competing answer tokens evolves across layers when a semantic-preserving paraphrase changes the model's response from correct to incorrect.

For each case, we compare the layer-wise log-probability difference between the first token of the correct answer and that of the incorrect answer generated under the paraphrased input. Formally, for an input question $q$ and each layer $\ell$, we compute:
\begin{equation}
    \Delta \log p^{(\ell)}
    =
    \log p^{(\ell)}(t_{\text{gold}} \mid q)
    -
    \log p^{(\ell)}(t_{\text{err}} \mid q),
\end{equation}
where $t_{\text{gold}}$ denotes the first token of the ground-truth answer and $t_{\text{err}}$ denotes the first token of the incorrect answer generated under the paraphrased input. A larger value indicates a stronger preference for the correct answer token, while a smaller value indicates a shift toward the incorrect one; accordingly, redder regions in the visualization indicate stronger correct-answer preference, and bluer regions indicate stronger incorrect-answer preference.
We focus on the first answer token to capture the model's initial factual preference and avoid confounding effects from downstream autoregressive dependencies.

Figure~\ref{fig:layer_heatmap_3} presents three representative factual inconsistency cases. Panels (a) and (b) correspond to two different paraphrases of the same original question, where the ground-truth answer is \textit{Greek}. Although the two paraphrases lead to different incorrect answers, namely \textit{film} and \textit{English}, they exhibit a similar pattern: the original question develops a stronger preference for the correct answer in later layers, while the paraphrased question shifts the late-layer preference toward the incorrect answer. Panel (c) shows a different type of error, where the correct answer \textit{spanish} is replaced by an \textit{unknown} response under the paraphrased input.

Across these cases, preference differences between original and paraphrased inputs remain relatively small in early and middle layers, but become more pronounced in later layers. This suggests that robustness failures under semantic-preserving paraphrases are associated with late-stage factual preference shifts, rather than a complete disruption of early lexical or semantic representations.

These observations align with prior layer-wise analyses showing that vocabulary projections can reveal evolving factual preferences across layers, and that later layers are closely related to factual prediction and answer calibration \cite{chuang2023dola,zhang2024sled}. We therefore use the heatmaps as diagnostic evidence that paraphrase-induced failures are accompanied by late-stage factual preference shifts, rather than as causal proof of the failure mechanism.

\end{document}